\pdfoutput=1
\documentclass{article}

\usepackage{arxiv}

\usepackage[utf8]{inputenc} % allow utf-8 input
\usepackage[T1]{fontenc}    % use 8-bit T1 fonts
\usepackage{hyperref}       % hyperlinks
\usepackage{url}            % simple URL typesetting
\usepackage{booktabs}       % professional-quality tables
\usepackage{amsfonts}       % blackboard math symbols
\usepackage{amsmath,amssymb}
\usepackage{nicefrac}       % compact symbols for 1/2, etc.
\usepackage{microtype}      % microtypography
\usepackage{graphicx}
\usepackage{multirow,array,tabularx,adjustbox}
\usepackage{xcolor}
\usepackage{caption}
\usepackage{natbib}
\usepackage{doi}
\newcolumntype{L}{>{\raggedright\arraybackslash}X}

\title{Statistical Machine Learning for Medicinal Plant Leaves Classification}

\author{ \href{https://orcid.org/0000-0002-6265-6481}{\includegraphics[scale=0.06]{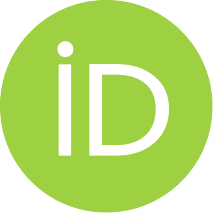}\hspace{1mm}Jayani P.~Gamage}\thanks{Corresponding author.} \\
	Centre for Crop and Disease Management (CCDM)\\
	Faculty of Science and Engineering\\
	Curtin University\\
	Bentley, WA, Australia \\
	\texttt{j.piyadigamage@curtin.edu.au} \\
	\And
	Thiyanga S.~Talagala \\
	Department of Statistics\\
	University of Sri Jayewardenepura\\
	Nugegoda, Sri Lanka \\
}

\renewcommand{\shorttitle}{Statistical Machine Learning for Medicinal Plant Leaves Classification}

\hypersetup{
colorlinks=true,linkcolor=blue,citecolor=blue,urlcolor=blue,
pdftitle={Statistical Machine Learning for Medicinal Plant Leaves Classification},
pdfsubject={cs.CV, cs.LG, stat.ML},
pdfauthor={Jayani P. Gamage, Thiyanga S. Talagala},
pdfkeywords={image processing, feature extraction, statistical machine learning, hierarchical classification, medicinal plants, leaf image classification, interpretable machine learning},
}

\begin{document}
\maketitle

\begin{abstract}
Medicinal plants are usually identified using practitioners’ experience, but reproducible, image based methods can complement this knowledge. We present MEDIPI (MEDIcinal Plant Identification), a statistical machine learning approach that classifies medicinal plant leaves from photographs. From each leaf image we extract 52 interpretable features covering shape, texture, color, and a new scagnostics category. We then classify leaves using a three level hierarchy: shape, then edge type, then species. We also release MedLEA, an open R package with 1,099 leaf images of 31 Sri Lankan medicinal plant species. On this dataset, a hierarchical Random Forest achieved the best overall accuracy (100\%) in our comparison of hierarchical and non-hierarchical approaches and eight algorithms: Random Forest, XGBoost, Gradient Boosting, linear discriminant analysis, k-nearest neighbor, support vector machine, probabilistic neural network, and artificial neural network. Adding texture, color, and scagnostic features to shape features raised shape level accuracy from 90\% to 97\%. Linear discriminant analysis plots show how the trained model separates leaf shapes. Accuracy dropped sharply (23–50\%) when training and test data came from different datasets. Therefore, generalization across datasets remains a challenge.
\end{abstract}

\keywords{image processing \and feature extraction \and statistical machine learning \and hierarchical classification \and medicinal plants \and leaf image classification \and interpretable machine learning}

\section{Introduction}
\label{sec:intro}

Located in the tropics, Sri Lanka is home to many plant species with medicinal
properties. Generations of Sri Lankans have used them as herbal treatments, and the
traditional medicine system, which has been practiced for more than $3000$ years, is
still in use today \citep{PMID}. It includes Ayurveda, Unani, and Deshiya Chikitsa
\citep{article20}. Traditional treatments are used to manage many conditions,
including diseases with complicated causes such as diabetes, arthritis, and cancer
\citep{PMID,articleintro1}, and a wide range of plants is used for this purpose
\citep{article20,8675114}.

\begin{figure}[htbp]
  \centering
  \includegraphics[width=0.47\textwidth]{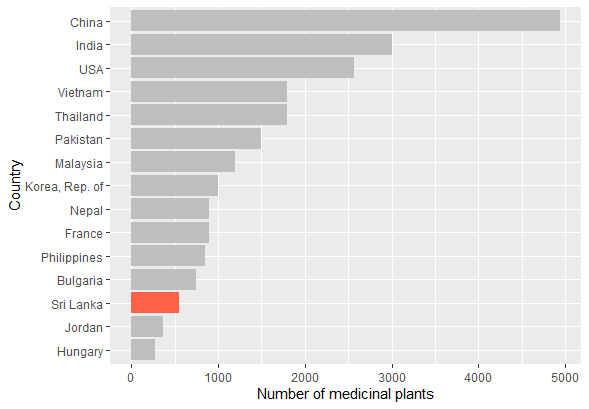}
  \hfill
  \includegraphics[width=0.47\textwidth]{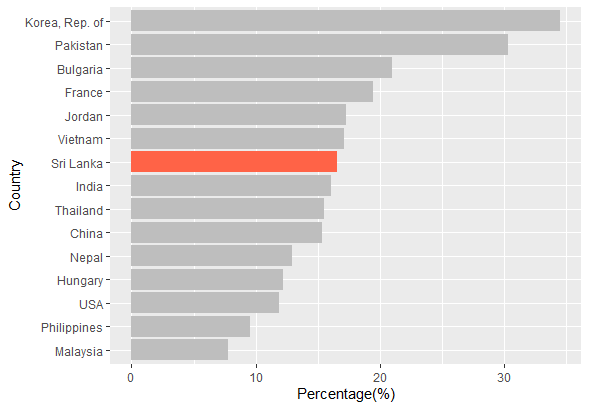}
  \caption{Sri Lanka is in the top 15 for medicinal plant diversity among countries with published estimates. (a) Number of medicinal plant species and (b) percentage of medicinal plants relative to total flora in 15 countries; Sri Lanka is highlighted. Data: \cite{schippmann2006impact}, Table~1.}
  \label{fig:1}
\end{figure}

Medicinal plants are important worldwide. An estimated $50,000$ to $70,000$ plant
species are used for medicinal purposes \citep{schippmann2006impact}, and the World
Health Organization's Global Traditional Medicine Strategy 2025--2034 stresses that
traditional medicines should be used safely and based on scientific evidence
\citep{who2025strategy}. Sri Lanka has about $550$ medicinal plant species out of a
total of $3,314$ plant species (16.6\%) \citep{schippmann2006impact}. Among the 15
countries with available data \citep{schippmann2006impact}, Sri Lanka ranks
$13^{\mathrm{th}}$ in the number of medicinal plant species and $7^{\mathrm{th}}$ in
the percentage of medicinal plants compared with its total flora
(Figure~\ref{fig:1}). Estimates of the number of medicinal plants in Sri Lanka differ
between sources; recent surveys report as many as 1432 species \citep{article12}.

\begin{figure}[htbp]
  \centering
  \includegraphics[width=0.47\textwidth]{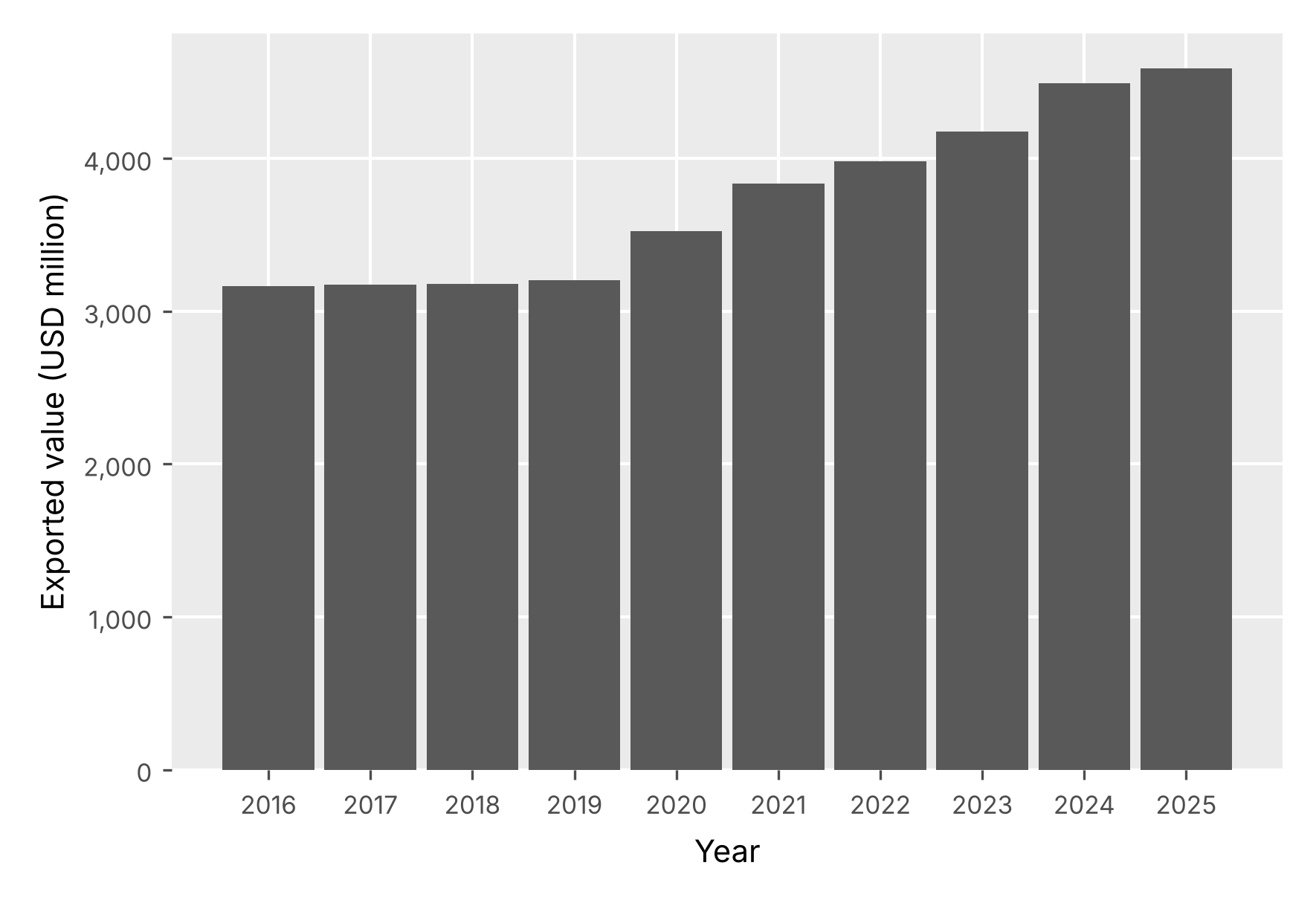}
  \hfill
  \includegraphics[width=0.47\textwidth]{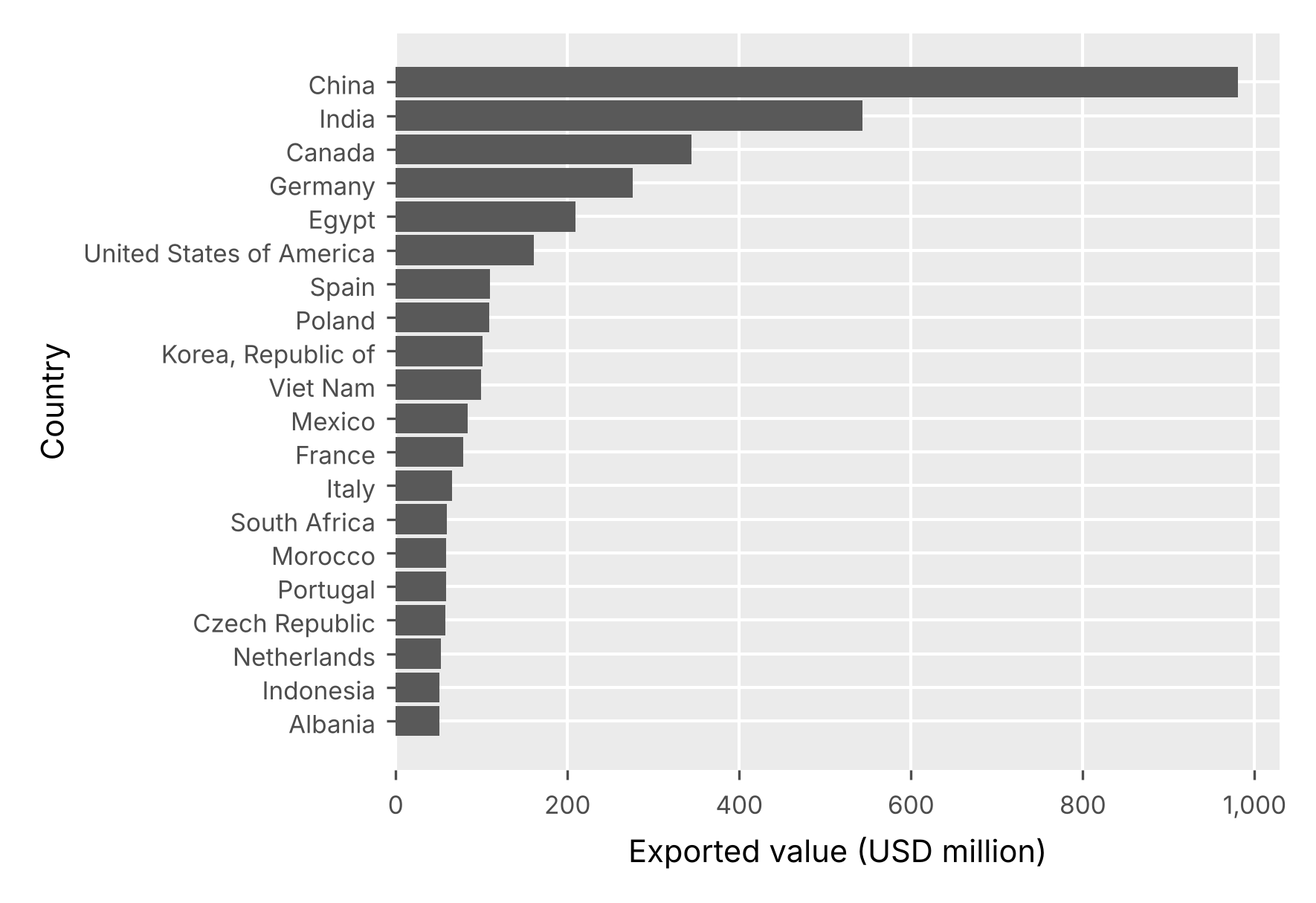}
  \caption{Global trade in medicinal plants is growing, but Sri Lanka's share is small. (a) Global export value of HS 1211 plants, 2016--2025 (2025 is provisional); (b) top 20 exporting economies of HS 1211 plants in 2024. Source: ITC Trade Map, \url{https://www.trademap.org}.}
  \label{fig:2}
\end{figure}

Sri Lanka also exports medicinal plants and plant materials. According to ITC Trade
Map data for HS code $1211$, which covers plants and plant parts used mainly in
perfumes, medicines, and insecticides \citep{trademap}, Sri Lanka's export value rose
from USD~3.9 million in 2016 to USD~17.1 million in 2025, more than four times the
2016 value. Over the same period, global exports rose from USD~3.2 billion to
USD~4.6 billion (Figure~\ref{fig:2}a). Even so, Sri Lanka's share of the global market
remains small, at about 0.4\%. In 2024, the five largest exporters were China, India,
Canada, Germany, and Egypt, while Sri Lanka ranked $41^{\mathrm{st}}$
(Figure~\ref{fig:2}b).

Despite this demand, there is no standard way to identify medicinal plants. Many
medicinal plants grow in backyards with little care, and some are even treated as
weeds \citep{PMID}. Most Sri Lankans know the traditional system and use the plants
growing near their homes without the advice of a practitioner, relying on knowledge
passed down by their ancestors \citep{article46}. Identifying plants by hand,
however, needs a lot of botanical expertise, and it is costly, slow, and difficult
for the general public. Misidentification also matters for conservation. Nearly
15{,}000 medicinal plant species are threatened worldwide, and 280 plant species are
threatened in Sri Lanka. Of the medicinal plants in Sri Lanka, 100--200 species are
threatened \citep{article12}, including \textit{Abarema begimena}, the Ashoka tree
(\textit{Saraca asoca}), the Beautiful Leaf (\textit{Calophyllum trapezifolium}), and
\textit{Aglaia apiocarpa}. The IUCN classes threatened plants as Critically
Endangered, Endangered, or Vulnerable.

Automatic identification from images is a promising solution, but existing work has
several gaps. Most algorithms learn from images, so image quality strongly affects
performance. Researchers therefore use the built-in cameras of mobile devices
\citep{articlee} or special cameras and scanners, and there are rules for taking the
photographs, such as a single leaf, good lighting, no shadows, and a fixed
line of sight angle \citep{8675114}. Most studies use secondary datasets such as
Flavia and Swedish \citep{8675114,articlee}. These datasets contain few plant species
\citep{8675114,articlee}, so researchers often collect their own images, and no
database of Sri Lankan medicinal plant leaves was available.

Images of leaves, flowers, bark, and fruit \citep{articlee} have all been used to train
classification models. Leaves carry many useful features and are easy to collect, so
most researchers use them, and most work focuses on shape features \citep{8675114}.
Shape features alone, however, are not enough to train a reliable model, which raises
the question of which features matter most for classifying plant species. Existing
algorithms are mostly built on CNN, ANN, PNN, and KNN architectures. These models
need a lot of memory and computing power, and although they can predict well, they
are hard to interpret. We address these gaps with an image feature-based statistical
machine learning algorithm.

Our main objective is to develop an automatic algorithm that classifies medicinal
plants from leaf images using statistical machine learning. To reach it, we pursue
four supporting objectives. First, we build and share a repository of medicinal plant
leaf images from Sri Lanka. Gathering leaf samples is slow, so sharing the database
gives other researchers data to train and test their own algorithms. Second, we
identify features that are useful for classifying medicinal plants from leaf images,
which contain a large and diverse set of features such as shape, veins, edges, and
apices, and we develop an algorithm to extract and quantify them. Third, we use
high-dimensional visualization to show what happens inside the trained algorithm.
Fourth, we release the algorithm as open-source software for identifying Sri Lankan
medicinal plants from leaf images.

This research can help avoid misidentification of medicinal plants in Sri Lanka,
which supports conservation and ecological efforts. Leaves are easy to collect
without damaging the plant, so developing and using the algorithm does no harm to
the plants. The algorithm is also hierarchical: users can follow its first two
levels even if they do not know the exact species name, which may reduce
misidentification and computation time.

The remainder of this paper is organized as follows. Section~\ref{sec:methodology}
describes the materials and methods, including the datasets, the image processing
and feature extraction pipeline, the hierarchical classification framework, and an
exploratory analysis of leaf morphology in the Ruhuna dataset.
Section~\ref{sec:results} presents the results: the two classification experiments, a
comparison with benchmark algorithms, visualization of the trained model using linear
discriminant analysis, and feature importance at each level of the hierarchy.
Section~\ref{sec:discussion} discusses the findings and their limitations, and
Section~\ref{sec:conclusions} concludes and outlines future work.

\section{Materials and Methods}
\label{sec:methodology}

\subsection{Overview of the Algorithm}

In this section, we provide a general overview of the methodology used to develop
our classification algorithm. The classification algorithm we introduce contains
two main phases: (i) the offline phase, and (ii) the online phase.

\begin{figure}[htbp]
  \centering
  \includegraphics[width=0.6\textwidth]{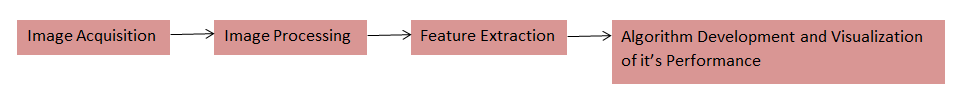}
  \caption{Workflow of the offline phase of the algorithm.}
  \label{fig:3}
\end{figure}

As shown in Figure~\ref{fig:3}, the workflow of the offline phase of the algorithm
contains four main steps:
\begin{enumerate}
  \item Image acquisition
  \item Image processing
  \item Feature extraction
  \item Algorithm development and visualization of algorithm performance
\end{enumerate}

\begin{figure}[htbp]
  \centering
  \includegraphics[width=0.85\textwidth]{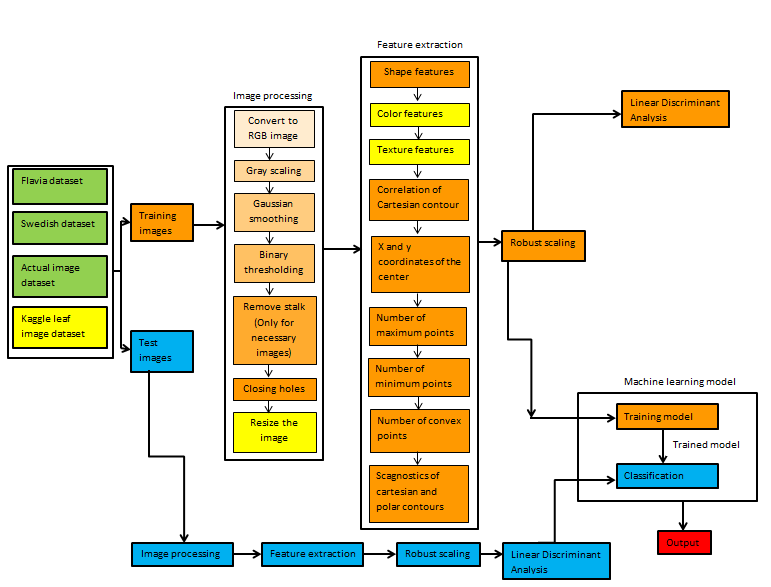}
  \caption{Methodology diagram.}
  \label{fig:4}
\end{figure}

Figure~\ref{fig:4} shows the overview of the methodology followed. The online
phase of the study is colored orange and the offline phase is colored blue.
Firstly, we acquire leaf images from existing datasets and from the leaf image
dataset collected by ourselves. Each leaf image dataset is then divided into
training and test images: the training set contains 80\% of images and the test
set contains 20\% of images from each dataset. We use four datasets to build and
evaluate the algorithm; a summary is given in Table~\ref{tab:1}.

\begin{table}[htbp]
\centering
\caption{Summary of datasets used in the algorithm.}
\label{tab:1}
\begin{tabular}{lcc}
\toprule
Dataset & Image format & Total number of leaf images \\
\midrule
Actual Leaf Image Dataset (MedLEA) & Color  & 1099 \\
Flavia dataset                     & Color  & 1907 \\
Swedish Leaf Image Dataset         & Color  & 1125  \\
Kaggle Leaf Image Dataset          & Binary & 1584 \\
\bottomrule
\end{tabular}
\end{table}

The next step is image processing. As shown in Figure~\ref{fig:4}, the main image
processing steps are: convert to RGB image, gray scaling, Gaussian smoothing,
binary thresholding, remove stalk, close holes, and resize image. Since the
Kaggle leaf image dataset contains only binary images, resizing is the only
applicable step. Removing the stalk and closing holes apply only to datasets
containing leaf images with a stalk or with holes (e.g., diseased leaves). After
image processing, images are ready for feature extraction. There are four
classes of features: (i) shape features, (ii) color features, (iii) texture
features, and (iv) scagnostics features of Cartesian and polar coordinates. In
this research we also introduce several new features: correlation of Cartesian
contour, $x$ and $y$ coordinates of the contour, number of minimum and maximum
points, and number of convex points. The resulting dataset contains all features
along with the leaf image id; note that the Kaggle leaf image dataset does not
have color and texture features. Robust scaling \citep{brownlee2020scale} is applied
to scale the data. To visualize the labeled feature dataset, linear
discriminant analysis (LDA) \citep{tharwat2017linear} is used. Our algorithm operates as a hierarchical classification system: leaves are
first classified according to shape (diamond, simple round, round, needle, or
heart shape), then according to edge type, and finally the bottom level classifies
plant species. Before training the model, all leaves are labeled according to
shape type, edge type, leaf arrangement, apex type, base type, etc. These labels
were identified by exploring the ``Ayurvedic Medicinal Plants of Sri Lanka''
medicinal leaf repository maintained by Barberyn Ayurveda Resort and the
University of Ruhuna. The information gathered from this repository is made
available through our R package \texttt{MedLEA}. The next step is to train the
model on the training dataset using machine learning techniques: a multi-class
supervised classification problem and use the trained model to predict labels
in the test dataset.

\begin{figure}[htbp]
  \centering
  \includegraphics[width=0.75\textwidth]{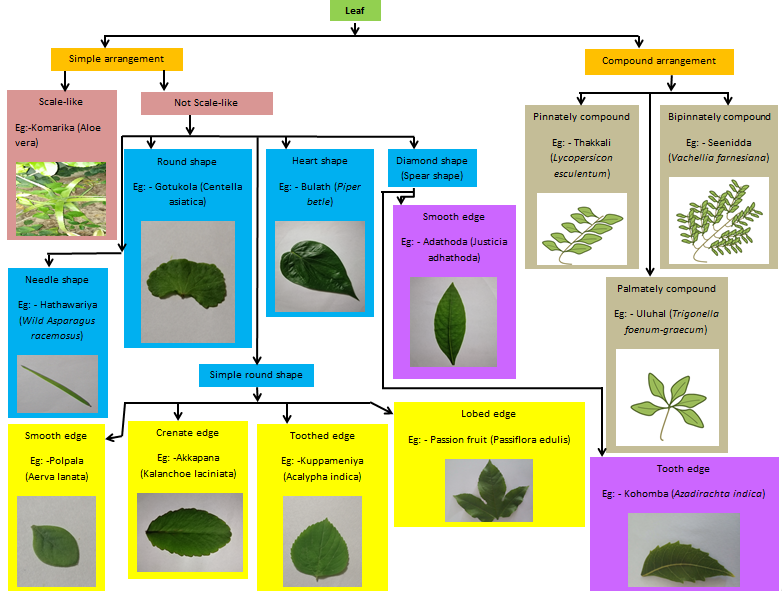}
  \caption{Classification hierarchy.}
  \label{fig:5}
\end{figure}

Our medicinal plant classification algorithm is defined as
MEDIPI: \textbf{MEDI}cinal \textbf{P}lant \textbf{I}dentification. MEDIPI is
divided into an offline phase and an online phase. The classification algorithm
is trained in the offline phase; in the online phase, the pre-trained
classification model is used for real-time leaf image classification for general
users.

\subsection{Data}

We use four datasets: one primary dataset and three secondary datasets. The
primary dataset is named MedLEA; the secondary datasets are Flavia, Swedish, and Kaggle (Table~\ref{tab:2}).

\begin{table}[htbp]
\centering
\caption{Summary of secondary datasets.}
\label{tab:2}
\small
\resizebox{\textwidth}{!}{%
\begin{tabular}{lccccp{3.7cm}}
\toprule
Dataset & Image format & No.\ species & Images per species & No.\ leaf images & Remarks \\
\midrule
Flavia   & Color  & 32 & 50--77 & 1907 & Scanners and digital cameras used to acquire leaf images on a plain background; isolated leaf images contain blades only, without petiole. Collected in China. \\
Swedish  & Color  & 15 & 75     & 1125 & Isolated leaf scans on a plain background. Collected in Sweden. \\
Kaggle   & Binary & 99 & 16     & 1584 & Collected in the United Kingdom. \\
\bottomrule
\end{tabular}
}
\end{table}

\begin{figure}[htbp]
  \centering
  \includegraphics[width=0.85\textwidth]{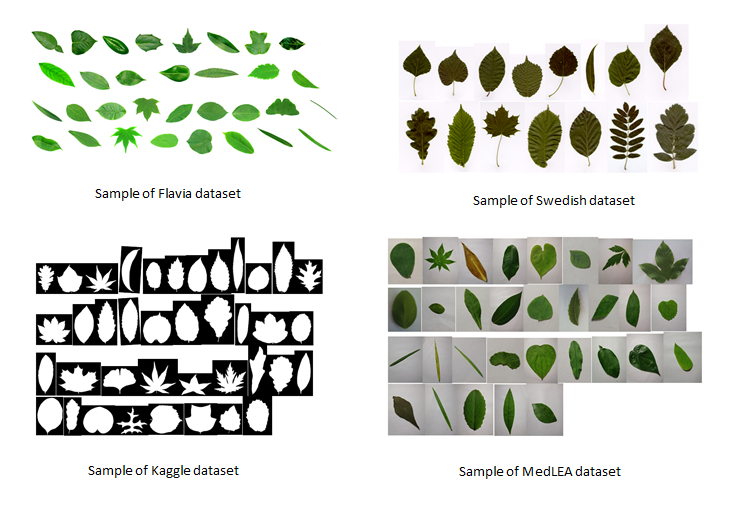}
  \caption{Sample leaf images from the Flavia, Swedish, Kaggle, and MedLEA datasets.}
  \label{fig:6}
\end{figure}

\subsubsection{MedLEA}

\begin{figure}[htbp]
  \centering
  \includegraphics[width=0.85\textwidth]{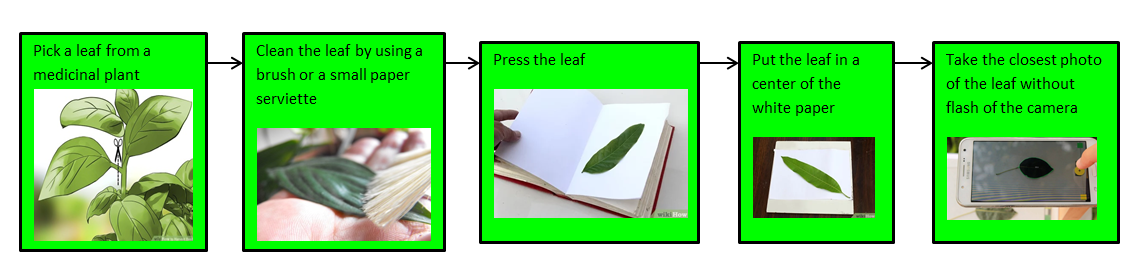}
  \caption{Image collection process of medicinal plants in Sri Lanka.}
  \label{fig:7}
\end{figure}

The image collection process contains five main steps, shown in
Figure~\ref{fig:7}. This approach is simple and can be followed without any
expert knowledge.

\begin{figure}[htbp]
  \centering
  \includegraphics[width=0.7\textwidth]{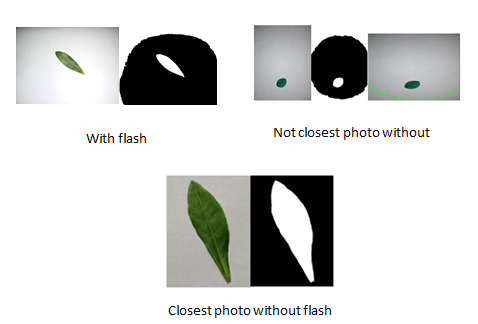}
  \caption{Effect of flash and distance on binary conversion quality.}
  \label{fig:8}
\end{figure}

First, a plant is selected and a leaf is picked using scissors, without petiole,
in the morning when the leaf looks fresh. The leaf is then cleaned with a small
brush or paper serviette to remove water bubbles, soil, or mud. In some cases,
where the leaf rounds from the apex, base, or margin and cannot lie flat, the leaf
is pressed for approximately 1--2 days before photographing, to avoid distorting
the extracted contour. The pressed leaf is then kept centered on a white sheet of
paper, since accurate binary conversion depends on the leaf being centered.
Finally, the closest photo is taken without flash (Figure~\ref{fig:8}): the
closest photo reduces computational load and unnecessary foreground region
\citep{goyal2018plant}, and avoiding flash removes shadow, which is important for
capturing the leaf shape correctly in the binary image. Photos are taken in
daylight to minimize the effect of light illumination.

Through this research, we establish a repository of
medicinal plant images in Sri Lanka. This repository is made available to the
public through the open-source R package \texttt{MedLEA}, available at
\url{https://CRAN.R-project.org/package=MedLEA}, for research reproducibility.

There are 1099 leaf images of 31 species, with 29--45 images per species of
medicinal plants in Sri Lanka. These leaves have simple arrangement: a single
leaf that is never divided into smaller leaflet units, always attached to a twig
by its stem or petiole. Leaf margins can be smooth, lobed, or toothed. Photos
were taken with a Huawei Nova 3i, with the closest photos captured on a white
background.

\subsection{Image Processing}

\begin{figure}[htbp]
  \centering
  \includegraphics[width=0.85\textwidth]{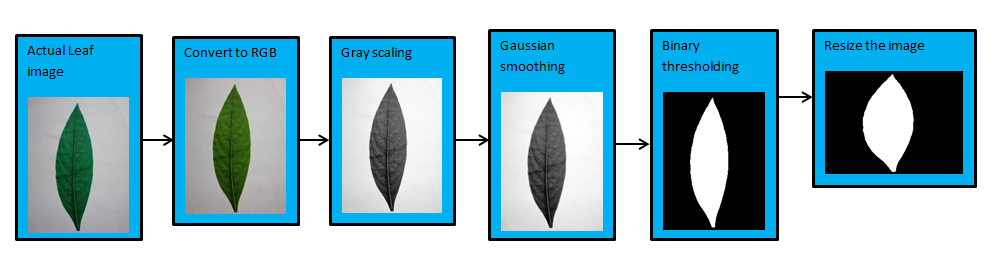}
  \caption{Image processing workflow.}
  \label{fig:9}
\end{figure}

Image processing plays a vital role in leaf image identification, reducing
noise, subtracting the background, and enhancing content for the identification
process \citep{goyal2018plant}. The workflow used to process images in this
paper is shown in Figure~\ref{fig:9} and includes seven main steps: (i)
converting BGR to RGB, (ii) gray scaling, (iii) Gaussian filtering, (iv) binary
thresholding, (v) removing stalk, (vi) closing holes, and (vii) image resizing.
Some steps apply only to specific images. For example, removing the stalk
applies only to leaf images that have one.

\begin{figure}[htbp]
  \centering
  \includegraphics[width=0.7\textwidth]{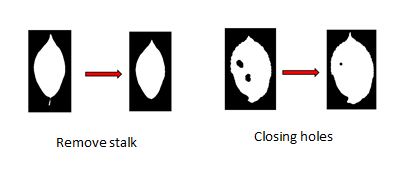}
  \caption{Binary image before and after removing the stalk and closing holes.}
  \label{fig:10}
\end{figure}

Feeding RGB images through gray scaling optimizes the contrast and intensity of
images by reducing dimensionality and complexity. Smoothing techniques are
applied to remove noise, and binary thresholding separates foreground from
background. Removal of the stalk and closing of holes in the foreground are
important for accurately capturing leaf shape (Figure~\ref{fig:10}). More details on the image processing pipeline are discussed in \cite{lakshika2021computer}.

\subsection{Feature Extraction}

The most crucial part of the algorithm is extracting distinctive leaf features
from the images. Much prior work focuses on neural network models such as CNNs
\citep{4458016,articlepl, herdiyeni2012mobile}, which are complex and
difficult to interpret. We instead introduce pre-calculated features that are
easy to interpret, generalize well, and are computationally efficient. We focus
on four types of leaf image features: shape, texture, color, and scagnostics
features, identifying 52 features in total. More details on these features are
discussed in \cite{lakshika2021computer}. 

\subsection{Classification Framework}
\label{sec:framework}

Our algorithm operates as a hierarchical classification system with three
levels. The first level classifies images according to shape; the second level
classifies according to edge type; the bottom level classifies plant species. If
species names were used directly rather than a hierarchical classification, the
resulting large number of class labels would be difficult to manage and would
create a severe class imbalance problem. To develop the hierarchical system we
explored the actual plant image repository ``Ayurvedic Medicinal Plants of Sri
Lanka'' (\url{http://www.instituteofayurveda.org/plants/}), which describes the
most commonly used medicinal plants in Ayurvedic practice in Sri Lanka. This
website was created as the result of a project implemented by Barberyn Ayurveda
resort and the University of Ruhuna, and was last updated on 11 May 2017. Our
pilot study was based on this database.

We investigated 471 medicinal leaves in this repository, recording the physical
appearances (morphological characteristics) of each leaf together with its
Sinhala (local) name, family name, and scientific name. There were 22 variables
in total, with a unique primary key ``Id'', of which 18 variables described
physical appearances such as leaf arrangement, shape, edge type, apex, and base.

\subsection{Exploratory Data Analysis of the Ruhuna Dataset}
\label{sec:eda}

In this section, we present exploratory data analysis to characterize the common
morphological features of leaves in the Ruhuna repository.

\begin{figure}[htbp]
  \centering
  \includegraphics[width=0.6\textwidth]{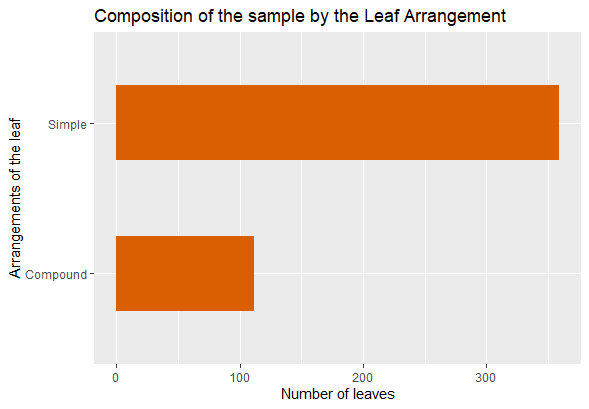}
  \caption{Composition of the Ruhuna dataset sample by leaf arrangement.}
  \label{fig:11}
\end{figure}

According to Figure~\ref{fig:11}, most leaves have simple arrangement; we
therefore restrict further analysis to leaves with simple arrangement. According
to Table~\ref{tab:4}, the most common leaf shapes are (i) diamond, (ii) simple
round, (iii) heart, (iv) needle, and (v) round, which we use as the first level
of the hierarchy. Most leaves have smooth edges; we identify four common edge
types: (i) smooth, (ii) toothed, (iii) lobed, and (iv) crenate, used as the
second level of the hierarchy. As shown in Table~\ref{tab:4}, most diamond,
heart, round, needle, and simple round-shaped leaves have smooth edges; all
needle-shaped leaves have smooth edges; there are no crenate-edged heart-shaped
leaves; and diamond, round, and simple round-shaped leaves span all four edge
types.

\begin{table}[htbp]
\centering
\caption{Cross tabulation of shape and edge type in the Ruhuna dataset.}
\label{tab:4}
\small
\resizebox{\textwidth}{!}{%
\begin{tabular}{llccccc}
\toprule
Shape label & & Smooth & Tooth & Lobed & Crenate & Row total (\%) \\
\midrule
\multirow{4}{*}{Diamond} & Count    & 167  & 27   & 22   & 1    & \multirow{4}{*}{217 (60.4)} \\
                          & Row \%   & 77.0 & 12.4 & 10.1 & 0.5  & \\
                          & Column \%& 66.0 & 44.3 & 55.0 & 20.0 & \\
                          & Total \% & 46.5 & 7.5  & 6.1  & 0.3  & \\
\midrule
\multirow{4}{*}{Heart}    & Count    & 26   & 14   & 6    & 0    & \multirow{4}{*}{46 (12.8)} \\
                          & Row \%   & 56.5 & 30.4 & 13.0 & 0.0  & \\
                          & Column \%& 10.3 & 23.0 & 15.0 & 0.0  & \\
                          & Total \% & 7.2  & 3.9  & 1.0  & 0.0  & \\
\midrule
\multirow{3}{*}{Needle}   & Count    & 21   & 0    & 0    & 0    & \multirow{3}{*}{21 (5.8)} \\
                          & Row \%   & 100  & 0.0  & 0.0  & 0.0  & \\
                          & Column \%& 8.3  & 0.0  & 0.0  & 0.0  & \\
\midrule
\multirow{3}{*}{Round}    & Count    & 5    & 3    & 1    & 1    & \multirow{3}{*}{10 (2.8)} \\
                          & Row \%   & 50   & 30   & 10   & 10   & \\
                          & Column \%& 2.0  & 4.9  & 2.5  & 20   & \\
\midrule
\multirow{3}{*}{Simple round} & Count    & 31   & 14   & 11   & 3    & \multirow{3}{*}{59 (16.4)} \\
                          & Row \%   & 52.5 & 23.7 & 18.6 & 5.1  & \\
                          & Column \%& 12.3 & 23.0 & 27.5 & 60   & \\
\midrule
\multirow{3}{*}{Scale-like} & Count    & 3   & 3    & 0    & 0    & \multirow{3}{*}{6 (1.7)} \\
                          & Row \%   & 50   & 50   & 0.0  & 0.0  & \\
                          & Column \%& 1.2  & 4.9  & 0.0  & 0.0  & \\
\midrule
Column total (\%) & & 253 (70.5) & 61 (17.0) & 40 (11.1) & 5 (1.4) & 359 (100) \\
\bottomrule
\end{tabular}
}
\end{table}

\begin{figure}[htbp]
  \centering
  \includegraphics[width=0.85\textwidth]{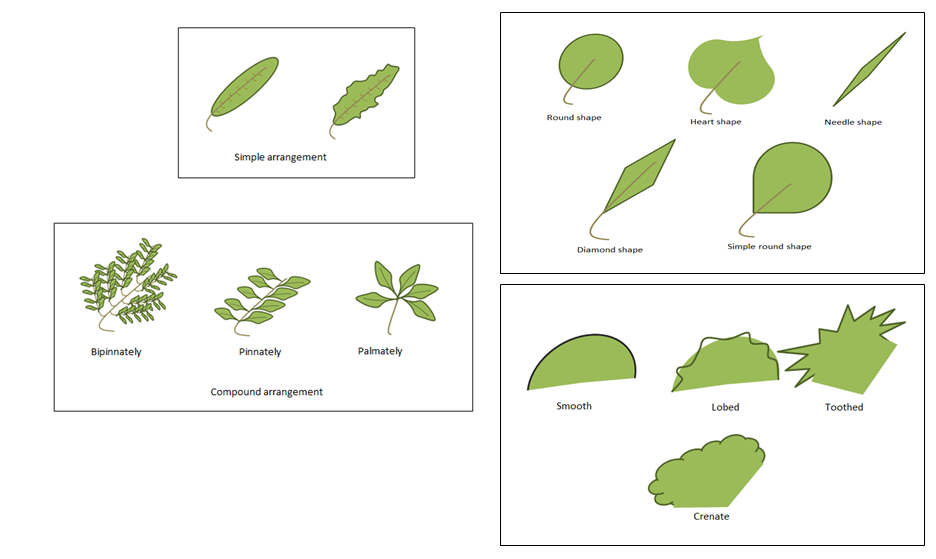}
  \caption{Common arrangements, shapes, and edge types.}
  \label{fig:12}
\end{figure}

\begin{figure}[htbp]
  \centering
  \includegraphics[width=0.95\textwidth]{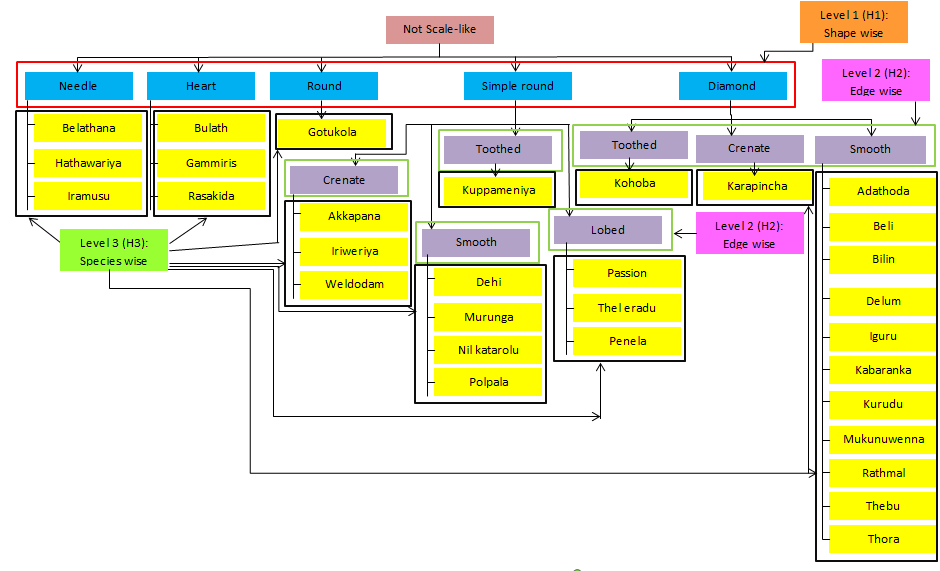}
  \caption{General hierarchical structure types.}
  \label{fig:13}
\end{figure}

By conducting this study we identified the common shapes and edge types of
medicinal leaves in Sri Lanka: five main shapes (diamond, heart, round, needle,
simple round) and four main edge types (smooth, toothed, lobed, crenate). This
informs the general hierarchical structure shown in Figure~\ref{fig:12} and
Figure~\ref{fig:13}.

\section{Results}
\label{sec:results}

\hypertarget{experiments}{%
\subsection{Experiments}
\label{experiments}}

\begin{figure}[htbp]
  \centering
  \includegraphics[width=75mm, height=60mm]{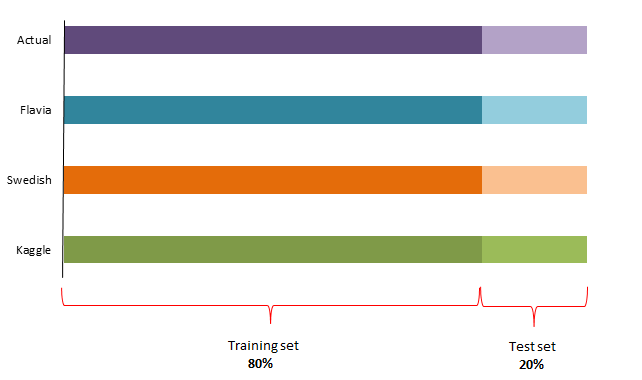}
  \includegraphics[width=75mm, height=60mm]{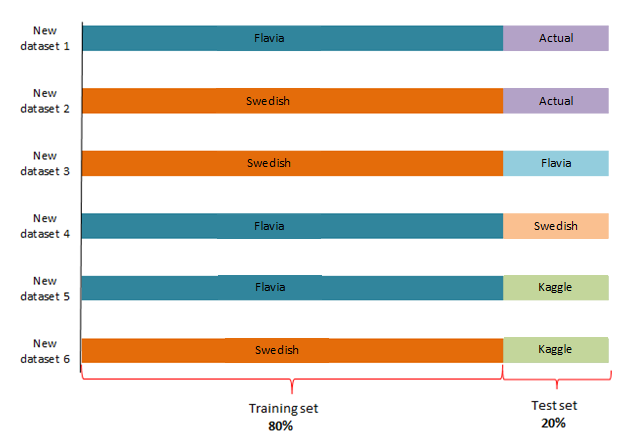}
  \caption{Proportion of training and test sets in Experiment~1 (a) and Experiment~2 (b).}
  \label{fig:23}
\end{figure}

We conducted two experiments, illustrated in Figure~\ref{fig:23}. Since most
studies in the literature consider shape features only, we designed the
experiments to test whether adding new feature categories improves accuracy,
arranging the experiments in two ways: (i) using all feature categories (shape,
color, texture, and scagnostic features), and (ii) using shape features only.

\hypertarget{experiment-1}{%
\subsection{Experiment 1}
\label{experiment-1}}

Experiment~1 asks how well the classifiers perform when the training and test sets
come from the \emph{same} dataset. We split each dataset into 80\% training and 20\%
test data, and we ran the experiment twice: first with all feature categories, and
then with shape features only. The Flavia, Swedish, and Kaggle datasets were used for
the first level of the hierarchy only (H1, shape). The Actual dataset (MedLEA) was
analyzed from top to bottom, so that we could record accuracy at every level: shape
(H1), edge type (H2), and species (H3).

We begin with the first level. Table~\ref{tab:5} shows shape-wise (H1) and overall
accuracy when all feature categories are used. The Kaggle dataset has no needle
shaped leaves, so that column is empty.

\begin{table}[htbp]
\centering
\caption{Shape-wise (H1) and overall classification accuracy, Experiment 1 -- all feature categories (training and test sets from the same dataset).}
\label{tab:5}
\small
\resizebox{\textwidth}{!}{%
\begin{tabular}{llcccccc}
\toprule
Dataset & Algorithm & Diamond & Simple Round & Needle & Heart & Round & Overall \\
\midrule
\multirow{3}{*}{Actual}  & Random Forest & 0.94 & \textbf{1.00} & 1.00 & \textbf{1.00} & 1.00 & \textbf{0.97} \\
                         & XGBoost       & 0.94 & 0.95 & \textbf{1.00} & 0.96 & 1.00 & 0.95 \\
                         & Gradient Boosting & \textbf{0.96} & 0.96 & \textbf{1.00} & \textbf{1.00} & 1.00 & \textbf{0.97} \\
\midrule
\multirow{3}{*}{Flavia}  & Random Forest & 0.98 & \textbf{0.97} & 0.96 & 0.97 & \textbf{1.00} & \textbf{0.98} \\
                         & XGBoost       & \textbf{0.99} & 0.95 & 0.93 & 0.98 & \textbf{1.00} & 0.97 \\
                         & Gradient Boosting & 0.98 & 0.96 & \textbf{1.00} & \textbf{1.00} & \textbf{1.00} & \textbf{0.98} \\
\midrule
\multirow{3}{*}{Swedish} & Random Forest & 0.99 & \textbf{1.00} & \textbf{1.00} & \textbf{0.95} & \textbf{1.00} & \textbf{0.98} \\
                         & XGBoost       & \textbf{1.00} & \textbf{1.00} & \textbf{1.00} & 0.88 & 0.92 & 0.97 \\
                         & Gradient Boosting & 0.99 & 0.98 & 0.96 & 0.86 & \textbf{1.00} & 0.96 \\
\midrule
\multirow{3}{*}{Kaggle}  & Random Forest & \textbf{0.76} & 0.69 & -- & 0.82 & \textbf{0.77} & \textbf{0.74} \\
                         & XGBoost       & 0.75 & 0.68 & -- & 0.70 & 0.70 & 0.71 \\
                         & Gradient Boosting & 0.74 & 0.68 & -- & \textbf{0.92} & 0.66 & 0.71 \\
\bottomrule
\end{tabular}
}
\end{table}

Random Forest classifies simple round, needle, round, and heart shaped leaves in the
Actual dataset with 100\% accuracy, and it ties with Gradient Boosting for the
highest overall accuracy (97\%). The same pattern holds in the other datasets: Random
Forest and Gradient Boosting tie on Flavia (98\%), and Random Forest is best on
Swedish (98\%) and Kaggle (74\%). Across all four datasets, the overall H1 accuracy of
Random Forest is equal to or higher than that of XGBoost and Gradient Boosting.

\begin{figure}[htbp]
  \centering
  \includegraphics[width=0.45\textwidth]{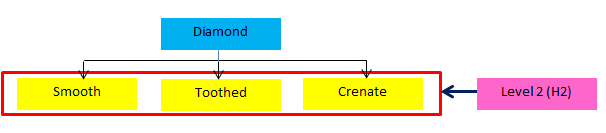}
  \hfill
  \includegraphics[width=0.45\textwidth]{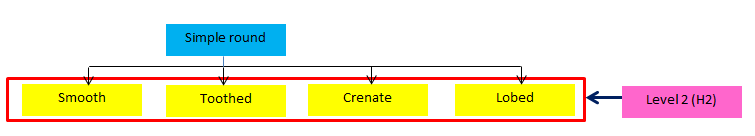}
  \caption{Level 2 hierarchy (H2) of diamond-shaped leaves (left) and simple round-shaped leaves (right).}
  \label{fig:1516}
\end{figure}

Because of this strong and consistent performance, we used Random Forest for the
lower levels of the hierarchy on the Actual dataset. Figure~\ref{fig:1516} shows the
second level (H2) for diamond and simple round leaves, and Table~\ref{tab:6} gives the
edge-wise accuracy.

\begin{table}[htbp]
\centering
\caption{Edge-wise (H2) and overall classification accuracy, Experiment 1 (training and test sets from the Actual dataset).}
\label{tab:6}
\small
\resizebox{\textwidth}{!}{%
\begin{tabular}{lccccccccc}
\toprule
& \multicolumn{4}{c}{Diamond} & \multicolumn{5}{c}{Simple Round} \\
\cmidrule(lr){2-5}\cmidrule(lr){6-10}
Dataset / Algorithm & Smooth & Toothed & Crenate & Overall & Smooth & Toothed & Lobed & Crenate & Overall \\
\midrule
Actual -- Random Forest & 1.00 & 1.00 & 0.40 & 0.97 & 0.94 & 1.00 & 1.00 & 1.00 & 0.97 \\
\bottomrule
\end{tabular}
}
\end{table}

At the edge level, Random Forest reaches 97\% overall accuracy for both diamond and
simple round leaves. Accuracy is 100\% for smooth and toothed diamond leaves, and for
toothed, lobed, and crenate simple round leaves. The weakest result is for crenate
diamond leaves (40\%), the only edge type that falls well below the rest.

\begin{figure}[htbp]
  \centering
  \includegraphics[width=0.45\textwidth]{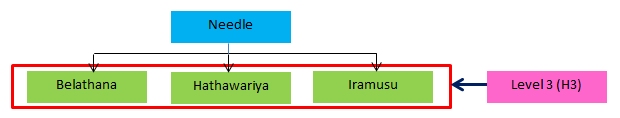}
  \hfill
  \includegraphics[width=0.45\textwidth]{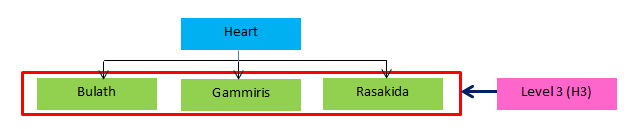}
  \caption{Level 3 hierarchy (H3) of needle-shaped leaves (left) and heart-shaped leaves (right).}
  \label{fig:1718}
\end{figure}

Next, we look at the species level (H3). Heart- and needle-shaped leaves have no edge
split, so their species are classified directly from shape (Figure~\ref{fig:1718} and
Table~\ref{tab:7}).

\begin{table}[htbp]
\centering
\caption{Species-wise (H3) and overall classification accuracy given shape (training and test sets from the Actual dataset).}
\label{tab:7}
\small
\resizebox{\textwidth}{!}{%
\begin{tabular}{lccccccccc}
\toprule
& \multicolumn{4}{c}{Needle} & \multicolumn{5}{c}{Heart Shape} \\
\cmidrule(lr){2-5}\cmidrule(lr){6-10}
Dataset / Algorithm & Belathana & Hathawariya & Iramusu & Overall & Bulath & Gammiris & Rasakida & & Overall \\
\midrule
Actual -- Random Forest & 1.00 & 1.00 & 1.00 & 1.00 & 1.00 & 1.00 & 1.00 & & 1.00 \\
\bottomrule
\end{tabular}
}
\end{table}

Random Forest classifies every species of needle- and heart-shaped leaves with 100\%
accuracy.

\begin{figure}[htbp]
  \centering
  \includegraphics[width=0.45\textwidth]{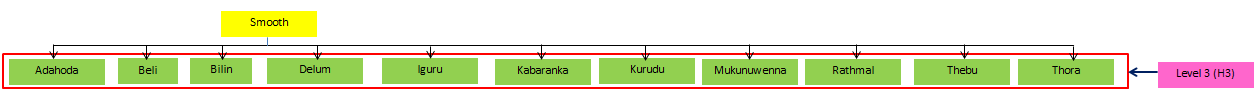}
  \includegraphics[width=0.45\textwidth]{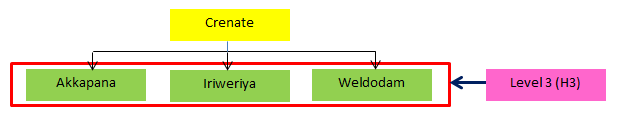}
  \includegraphics[width=0.45\textwidth]{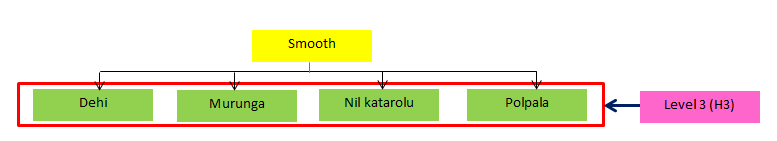}
  \includegraphics[width=0.45\textwidth]{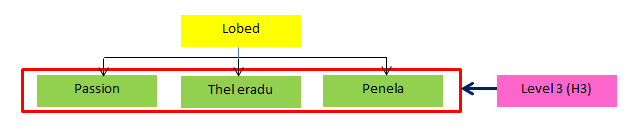}
  \caption{Level 3 hierarchy (H3) of diamond shaped smooth edged leaves, simple round shaped smooth edged leaves, simple round shaped crenate edged, and simple round shaped lobed edged leaves.}
  \label{fig:1922}
\end{figure}

For diamond and simple round leaves, species are classified within each edge type.
Figure~\ref{fig:1922} shows the hierarchy for these cases, and Table~\ref{tab:8}
gives the species-wise accuracy.

\begin{table}[!htbp]
\centering
\caption{Species-wise (H3) and overall classification accuracy
given the edge (training and test sets are from the Actual dataset)}
\label{tab:8}

\scriptsize
\setlength{\tabcolsep}{3pt}
\renewcommand{\arraystretch}{1.4}

\begin{adjustbox}{max width=\textwidth}
\begin{tabular}{l*{25}{c}}

\toprule

\multirow{3}{*}{\textbf{Dataset}}
& \multicolumn{12}{c}{\textbf{Diamond}}
& \multicolumn{13}{c}{\textbf{Simple Round}} \\

\cmidrule(lr){2-13}
\cmidrule(lr){14-26}

& \multicolumn{12}{c}{Smooth}
& \multicolumn{5}{c}{Smooth}
& \multicolumn{4}{c}{Lobed}
& \multicolumn{4}{c}{Crenate} \\

\cmidrule(lr){2-13}
\cmidrule(lr){14-18}
\cmidrule(lr){19-22}
\cmidrule(lr){23-26}

& \rotatebox{90}{Adathoda}
& \rotatebox{90}{Beli}
& \rotatebox{90}{Bilin}
& \rotatebox{90}{Delum}
& \rotatebox{90}{Iguru}
& \rotatebox{90}{Kabaranga}
& \rotatebox{90}{Kurudu}
& \rotatebox{90}{Mukunuwenna}
& \rotatebox{90}{Rathmal}
& \rotatebox{90}{Thebu}
& \rotatebox{90}{Thora}
& \rotatebox{90}{Overall Accuracy}

& \rotatebox{90}{Dehi}
& \rotatebox{90}{Murunga}
& \rotatebox{90}{Nil katarolu}
& \rotatebox{90}{Polpala}
& \rotatebox{90}{Overall Accuracy}

& \rotatebox{90}{Passion}
& \rotatebox{90}{Penela}
& \rotatebox{90}{Thel eradu}
& \rotatebox{90}{Overall Accuracy}

& \rotatebox{90}{Akkapana}
& \rotatebox{90}{Iriweriya}
& \rotatebox{90}{Weldodam}
& \rotatebox{90}{Overall Accuracy} \\

\midrule

\textbf{Actual}
& \multicolumn{25}{c}{} \\

Random Forest
& 1.00 & 1.00 & 1.00 & 0.50
& 1.00 & 0.43 & 1.00 & 0.88
& 0.50 & 1.00 & 0.88 & \textbf{0.87}

& 1.00 & 1.00 & 1.00 & 1.00
& \textbf{1.00}

& 1.00 & 1.00 & 1.00
& \textbf{1.00}

& 1.00 & 1.00 & 1.00
& \textbf{1.00} \\

\bottomrule

\end{tabular}
\end{adjustbox}
\end{table}

Random Forest achieves 100\% overall species-wise accuracy for simple round leaves of
every edge type. For smooth diamond leaves the overall accuracy is 87\%. Most species
are classified correctly, but three have noticeably lower accuracy: Kabaranga (43\%),
Delum (50\%), and Rathmal (50\%). These species may deserve a closer look.

Finally, we repeated the H1 experiment using shape features only. Table~\ref{tab:9}
shows the results.

\begin{table}[htbp]
\centering
\caption{Shape-wise (H1) and overall classification accuracy, Experiment 1 -- shape features only (training and test sets from the same datasets).}
\label{tab:9}
\small
\resizebox{\textwidth}{!}{%
\begin{tabular}{llcccccc}
\toprule
Training/Test set & Algorithm & Diamond & Simple Round & Needle & Heart & Round & Overall \\
\midrule
\multirow{3}{*}{Actual}  & Random Forest & 0.89 & 0.86 & \textbf{1.00} & \textbf{1.00} & 0.86 & 0.90 \\
                         & XGBoost       & 0.92 & \textbf{0.89} & 0.97 & 0.97 & \textbf{1.00} & \textbf{0.92} \\
                         & Gradient Boosting & \textbf{0.93} & 0.83 & \textbf{1.00} & 0.92 & 0.71 & 0.90 \\
\midrule
\multirow{3}{*}{Flavia}  & Random Forest & 0.93 & \textbf{1.00} & 0.91 & 0.97 & \textbf{1.00} & 0.96 \\
                         & XGBoost       & \textbf{0.96} & 0.96 & \textbf{1.00} & 0.96 & 0.98 & 0.96 \\
                         & Gradient Boosting & 0.94 & 0.98 & 0.95 & \textbf{1.00} & \textbf{1.00} & 0.96 \\
\midrule
\multirow{3}{*}{Swedish} & Random Forest & 0.96 & \textbf{1.00} & \textbf{1.00} & 0.85 & 0.94 & 0.95 \\
                         & XGBoost       & \textbf{0.99} & \textbf{1.00} & \textbf{1.00} & 0.82 & \textbf{1.00} & \textbf{0.97} \\
                         & Gradient Boosting & 0.95 & 0.98 & \textbf{1.00} & \textbf{1.00} & 0.92 & 0.96 \\
\bottomrule
\end{tabular}
}
\end{table}

With shape features only, XGBoost has the highest overall H1 accuracy on the Actual
(92\%) and Swedish (97\%) datasets, and all three algorithms tie on Flavia (96\%).
Comparing Table~\ref{tab:9} with Table~\ref{tab:5}, overall accuracy is lower when
only shape features are used. For Random Forest, accuracy drops from 97\% to 90\% on
the Actual dataset, from 98\% to 96\% on Flavia, and from 98\% to 95\% on Swedish.
Color, texture, and scagnostic features therefore add useful information beyond
shape.

\hypertarget{experiment-2}{%
\subsection{Experiment 2}
\label{experiment-2}}

Experiment~2 tests how well the classifiers generalize to data they have not seen
before. Here, the training and test sets come from \emph{different} datasets: each
model is trained on 80\% of one dataset and tested on 20\% of another (written as
Train $\to$ Test in the tables). As in Experiment~1, we ran the experiment with all
feature categories and then with shape features only. Only the first level of the
hierarchy (H1, shape) was evaluated.

Table~\ref{tab:10} shows shape-wise (H1) and overall accuracy when all feature
categories are used.

\begin{table}[htbp]
\centering
\caption{Shape-wise (H1) and overall classification accuracy, Experiment 2 -- all feature categories (training and test sets from different datasets).}
\label{tab:10}
\small
\resizebox{\textwidth}{!}{%
\begin{tabular}{llcccccc}
\toprule
Train $\to$ Test & Algorithm & Diamond & Simple Round & Needle & Heart & Round & Overall \\
\midrule
\multirow{3}{*}{Flavia $\to$ Actual}  & Random Forest & 0.70 & 0.28 & 0.23 & 0.04 & 0.10 & 0.41 \\
                         & XGBoost & 0.65 & 0.28 & \textbf{0.56} & 0.03 & \textbf{0.29} & \textbf{0.43} \\
                         & Gradient Boosting & 0.63 & \textbf{0.31} & 0.45 & 0.03 & 0.26 & 0.42 \\
\multirow{3}{*}{Swedish $\to$ Actual} & Random Forest & 0.38 & 0.22 & 0.06 & \textbf{0.12} & 0.39 & 0.26 \\
                         & XGBoost & 0.40 & 0.23 & 0.18 & 0.04 & 0.23 & 0.25 \\
                         & Gradient Boosting & \textbf{0.50} & \textbf{0.33} & 0.11 & \textbf{0.12} & \textbf{0.55} & \textbf{0.36} \\
\multirow{3}{*}{Swedish $\to$ Flavia} & Random Forest & 0.29 & \textbf{0.43} & 0.46 & 0.0 & 0.0 & 0.27 \\
                         & XGBoost & \textbf{0.42} & 0.34 & 0.40 & 0.01 & 0.01 & 0.31 \\
                         & Gradient Boosting & 0.40 & 0.37 & \textbf{0.53} & \textbf{0.12} & \textbf{0.27} & \textbf{0.37} \\
\multirow{3}{*}{Flavia $\to$ Swedish} & Random Forest & 0.57 & 0.72 & 0.03 & 0.0 & 0.01 & 0.39 \\
                         & XGBoost & \textbf{0.58} & 0.67 & \textbf{0.13} & 0.0 & \textbf{0.15} & \textbf{0.41} \\
                         & Gradient Boosting & 0.51 & \textbf{0.79} & 0.12 & 0.0 & 0.18 & \textbf{0.41} \\
\multirow{3}{*}{Flavia $\to$ Kaggle}  & Random Forest & \textbf{0.66} & 0.39 & 0.0 & 0.0 & \textbf{0.17} & \textbf{0.44} \\
                         & XGBoost & 0.56 & \textbf{0.48} & 0.0 & 0.0 & 0.21 & 0.43 \\
                         & Gradient Boosting & 0.51 & 0.47 & 0.0 & 0.0 & \textbf{0.25} & 0.42 \\
\multirow{3}{*}{Swedish $\to$ Kaggle} & Random Forest & 0.42 & 0.12 & 0.0 & 0.07 & 0.05 & 0.24 \\
                         & XGBoost & 0.35 & 0.12 & 0.0 & \textbf{0.10} & \textbf{0.14} & 0.23 \\
                         & Gradient Boosting & \textbf{0.44} & \textbf{0.14} & -- & \textbf{0.10} & 0.13 & \textbf{0.27} \\
\bottomrule
\end{tabular}
}
\end{table}

Accuracy drops sharply compared with Experiment~1. Overall accuracy ranges from 23\%
to 44\%, and no algorithm is consistently best. Random Forest is highest only for
Flavia $\to$ Kaggle (44\%), XGBoost for Flavia $\to$ Actual (43\%), and Gradient
Boosting for Swedish $\to$ Actual, Swedish $\to$ Flavia, and Swedish $\to$ Kaggle
(27\% to 37\%). XGBoost and Gradient Boosting tie for Flavia $\to$ Swedish (41\%).
Performance also varies by shape: heart-shaped leaves are rarely classified
correctly (at most 12\%).

Table~\ref{tab:11} shows the same experiment using shape features only. This table
covers four of the six dataset combinations; the two Kaggle combinations are not
included.

\begin{table}[htbp]
\centering
\caption{Shape-wise (H1) and overall classification accuracy, Experiment 2 -- shape features only (training and test sets from different datasets).}
\label{tab:11}
\small
\resizebox{\textwidth}{!}{%
\begin{tabular}{llcccccc}
\toprule
Train $\to$ Test & Algorithm & Diamond & Simple Round & Needle & Heart & Round & Overall \\
\midrule
\multirow{3}{*}{Flavia $\to$ Actual}  & Random Forest & 0.57 & 0.29 & 0.81 & 0.04 & 0.13 & 0.42 \\
                         & XGBoost & \textbf{0.65} & \textbf{0.37} & \textbf{0.85} & 0.03 & \textbf{0.48} & \textbf{0.50} \\
                         & Gradient Boosting & 0.55 & 0.36 & \textbf{0.95} & \textbf{0.05} & \textbf{0.58} & 0.49 \\
\multirow{3}{*}{Swedish $\to$ Actual} & Random Forest & 0.37 & \textbf{0.31} & \textbf{0.09} & \textbf{0.10} & 0.14 & \textbf{0.24} \\
                         & XGBoost & 0.38 & \textbf{0.35} & 0.0 & 0.05 & 0.08 & 0.23 \\
                         & Gradient Boosting & \textbf{0.53} & 0.27 & 0.03 & 0.08 & \textbf{0.39} & \textbf{0.31} \\
\multirow{3}{*}{Swedish $\to$ Flavia} & Random Forest & 0.29 & \textbf{0.45} & 0.38 & 0.0 & 0.0 & 0.27 \\
                         & XGBoost & \textbf{0.46} & 0.32 & \textbf{0.41} & 0.0 & \textbf{0.06} & 0.33 \\
                         & Gradient Boosting & \textbf{0.52} & 0.36 & 0.27 & \textbf{0.13} & \textbf{0.20} & \textbf{0.38} \\
\multirow{3}{*}{Flavia $\to$ Swedish} & Random Forest & \textbf{0.46} & \textbf{0.88} & 0.15 & 0.0 & \textbf{0.01} & \textbf{0.39} \\
                         & XGBoost & \textbf{0.47} & 0.72 & \textbf{0.16} & 0.0 & \textbf{0.01} & 0.36 \\
                         & Gradient Boosting & 0.39 & 0.84 & \textbf{0.16} & 0.0 & \textbf{0.01} & 0.36 \\
\bottomrule
\end{tabular}
}
\end{table}

The pattern is again mixed. Overall accuracy ranges from 23\% to 50\%. XGBoost is
highest for Flavia $\to$ Actual (50\%), Gradient Boosting for Swedish $\to$ Actual
(31\%) and Swedish $\to$ Flavia (38\%), and Random Forest for Flavia $\to$ Swedish
(39\%). Some individual shapes are classified well: needle-shaped leaves reach 81\%
to 95\% for Flavia $\to$ Actual, and simple round leaves reach up to 88\% for
Flavia $\to$ Swedish. Heart-shaped leaves remain very hard to classify (at most 13\%).

Taken together, the two tables show that the best algorithm depends on the dataset
pair. Gradient Boosting has the highest overall accuracy most often (in 5 of the 10
combinations across Tables~\ref{tab:10} and~\ref{tab:11}), but the differences between
algorithms are small compared with the drop from Experiment~1. Overall accuracy was
71\% to 98\% when training and test data came from the same dataset
(Tables~\ref{tab:5} and~\ref{tab:9}), and only 23\% to 50\% when they came from
different datasets. Generalizing across independently collected datasets is
therefore much harder than generalizing within one.

\hypertarget{algorithm-performance-and-comparison-with-benchmarks}{%
\subsection{Algorithm Performance \& Comparison with Benchmarks}
\label{algorithm-performance-and-comparison-with-benchmarks}}

This section asks two questions: does the hierarchical approach classify better than
a single, non-hierarchical classifier, and how does our best model compare with
other well-known algorithms? We used the Actual (MedLEA) image dataset for both.

Random Forest, XGBoost, and Gradient Boosting were tested with both the hierarchical
and the non-hierarchical approach. Five further algorithms were tested with the
non-hierarchical approach only: Linear Discriminant Analysis (LDA), $k$-Nearest
Neighbor (KNN), Support Vector Machine (SVM), Probabilistic Neural Network (PNN),
and Artificial Neural Network (ANN). We compared the algorithms using three
measures: overall accuracy, macro average accuracy, and weighted average accuracy.
We ranked the algorithms on each measure and picked the best one using the average
of the three ranks. Table~\ref{tab:12} shows the results.

\begin{table}[htbp]
\centering
\caption{Comparison of overall classification accuracy, macro average accuracy, and weighted average accuracy, with ranks, hierarchical vs.\ non-hierarchical approaches.}
\label{tab:12}
\small
\resizebox{\textwidth}{!}{%
\begin{tabular}{llcccccc}
\toprule
Case & Algorithm & \multicolumn{2}{c}{Overall} & \multicolumn{2}{c}{Macro avg.} & \multicolumn{2}{c}{Weighted avg.} \\
 & & Acc.\ & Rank & Acc.\ & Rank & Acc.\ & Rank \\
\midrule
\multirow{3}{*}{Hierarchical} & Random Forest & \textbf{1.00} & 1 & \textbf{0.99} & 1 & \textbf{1.00} & 1 \\
 & XGBoost & 0.99 & 2 & 0.99 & 1 & 0.99 & 2 \\
 & Gradient Boosting & 0.99 & 2 & 0.99 & 1 & 0.99 & 2 \\
\midrule
\multirow{8}{*}{Non-hierarchical} & Random Forest & 0.99 & 2 & 0.99 & 1 & 0.99 & 2 \\
 & XGBoost & 0.98 & 5 & 0.98 & 5 & 0.98 & 5 \\
 & Gradient Boosting & 0.98 & 5 & 0.98 & 5 & 0.98 & 5 \\
 & LDA & 0.98 & 5 & 0.98 & 5 & 0.98 & 5 \\
 & KNN & 0.84 & 9 & 0.81 & 9 & 0.84 & 9 \\
 & SVM & 0.47 & 10 & 0.40 & 11 & 0.42 & 11 \\
 & PNN & 0.47 & 10 & 0.44 & 10 & 0.45 & 10 \\
 & ANN & 0.98 & 5 & 0.98 & 5 & 0.98 & 5 \\
\bottomrule
\end{tabular}
}
\end{table}

The hierarchical Random Forest is the best model. It has the highest overall and
weighted average accuracy (1.00), and it ranks first on all three measures, so its
average rank is 1. On macro-average accuracy (0.99), it shares first place with the
hierarchical XGBoost and Gradient Boosting models and with the non-hierarchical
Random Forest.

The hierarchical approach gives a small but consistent gain. For each of the three
algorithms tested both ways, the hierarchical version scores one percentage point
higher than the non-hierarchical version on overall accuracy (Random Forest 1.00
vs.\ 0.99, XGBoost and Gradient Boosting 0.99 vs.\ 0.98). The benchmark algorithms
fall into three groups. LDA and ANN match the non-hierarchical boosting models at
0.98. KNN is lower (0.84). SVM and PNN perform worst, with overall accuracy of only
0.47.

\hypertarget{visualization-of-algorithm-performance}{%
\subsection{Visualization of Algorithm Performance}
\label{visualization-of-algorithm-performance}}

\begin{figure}[htbp]
  \centering
  \includegraphics[width=0.48\textwidth]{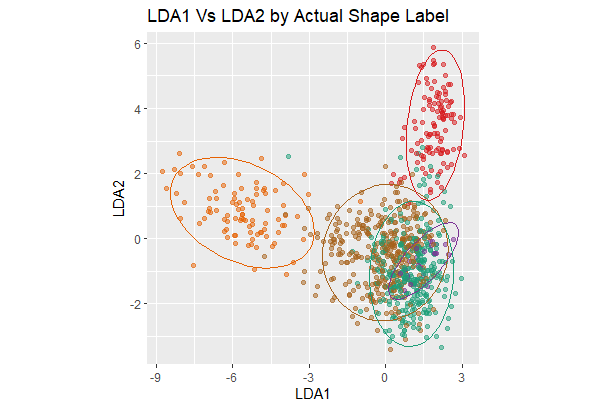}\hfill
  \includegraphics[width=0.48\textwidth]{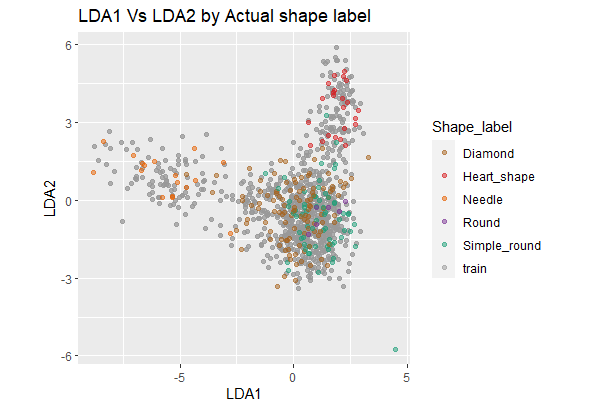}

  \medskip
  \includegraphics[width=0.48\textwidth]{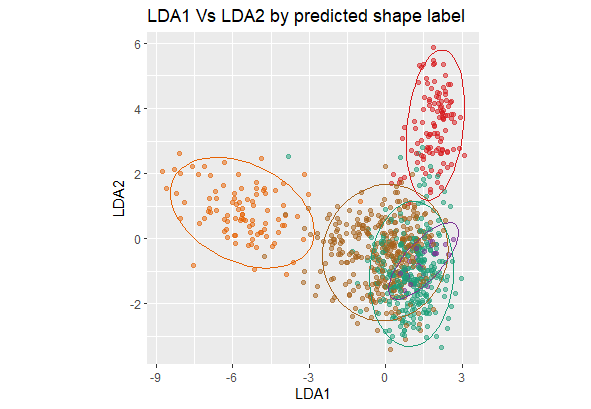}\hfill
  \includegraphics[width=0.48\textwidth]{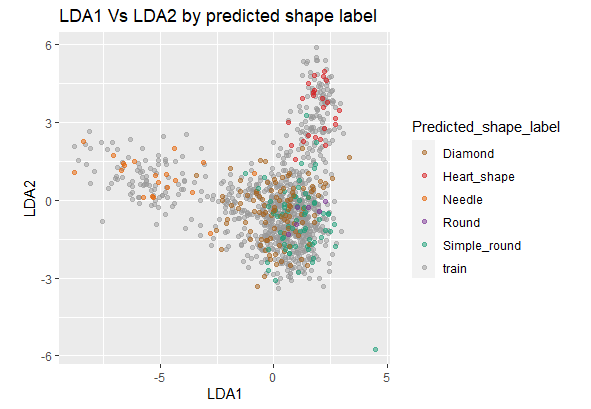}
  \caption{Needle- and heart-shaped leaves separate cleanly on the first two LDA axes. Leaves of the Actual dataset plotted on LDA1 and LDA2: (a) actual shape labels, training set; (b) actual labels, training + test sets; (c) labels predicted by Random Forest, training set; (d) predicted labels, training + test sets.}
  \label{fig:lda12}
\end{figure}

\begin{figure}[htbp]
  \centering
  \includegraphics[width=0.48\textwidth]{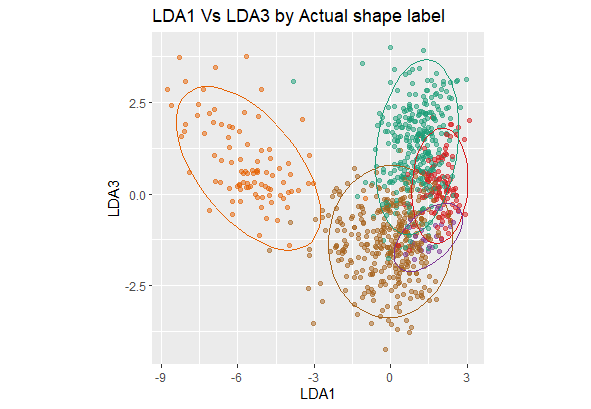}\hfill
  \includegraphics[width=0.48\textwidth]{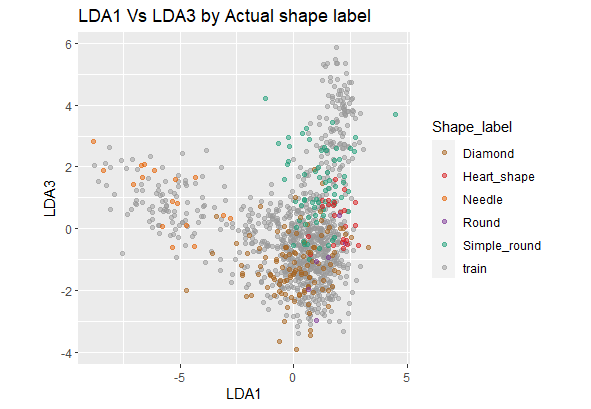}

  \medskip
  \includegraphics[width=0.48\textwidth]{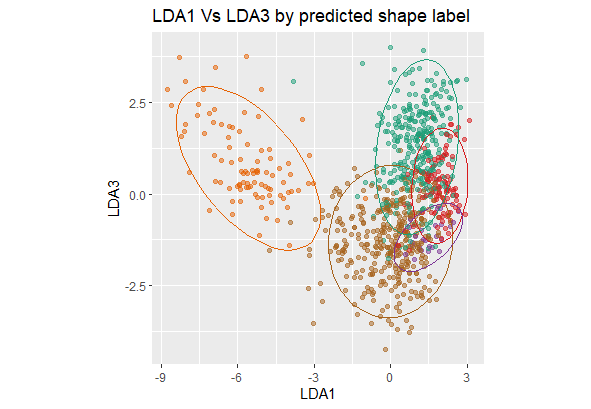}\hfill
  \includegraphics[width=0.48\textwidth]{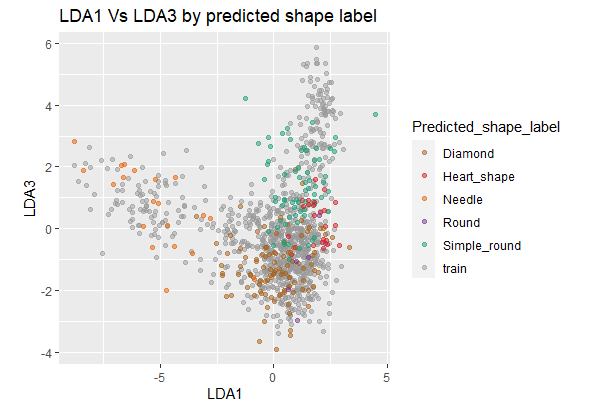}
  \caption{On LDA1 and LDA3, only needle-shaped leaves are clearly separated. Leaves of the Actual dataset plotted on LDA1 and LDA3: (a) actual labels, training set; (b) actual labels, training + test sets; (c) predicted labels, training set; (d) predicted labels, training + test sets.}
  \label{fig:lda13}
\end{figure}

\begin{figure}[htbp]
  \centering
  \includegraphics[width=0.48\textwidth]{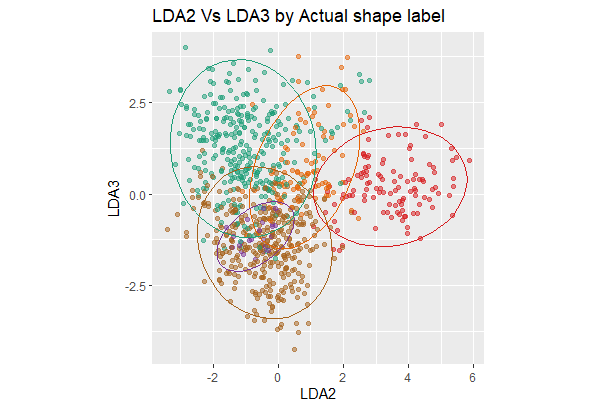}\hfill
  \includegraphics[width=0.48\textwidth]{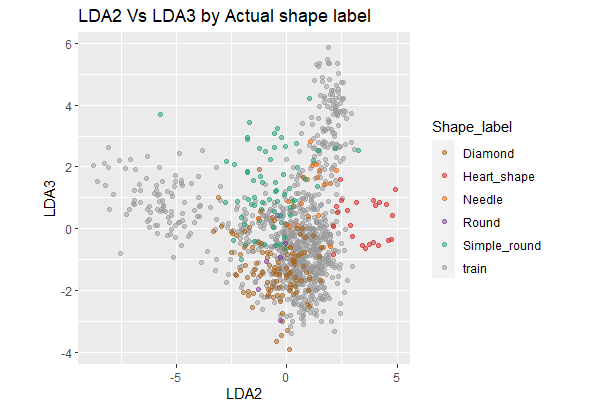}

  \medskip
  \includegraphics[width=0.48\textwidth]{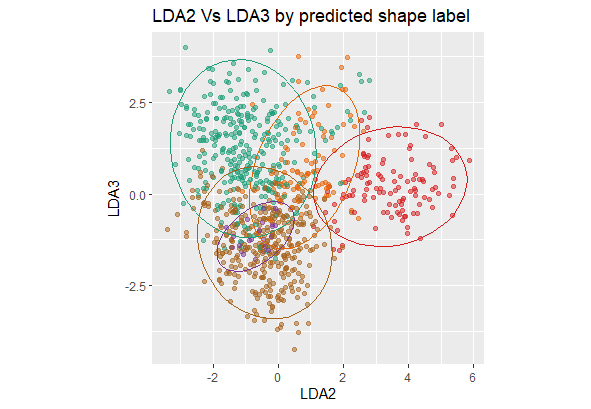}\hfill
  \includegraphics[width=0.48\textwidth]{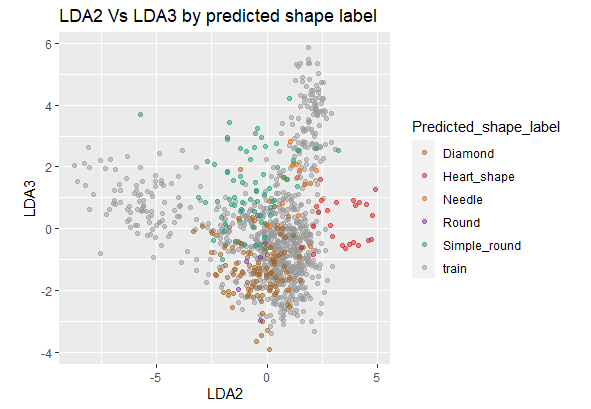}
  \caption{On LDA2 and LDA3, all shape labels overlap. Leaves of the Actual dataset plotted on LDA2 and LDA3: (a) actual labels, training set; (b) actual labels, training + test sets; (c) predicted labels, training set; (d) predicted labels, training + test sets.}
  \label{fig:lda23}
\end{figure}

\begin{figure}[htbp]
  \centering
  \includegraphics[width=0.95\textwidth]{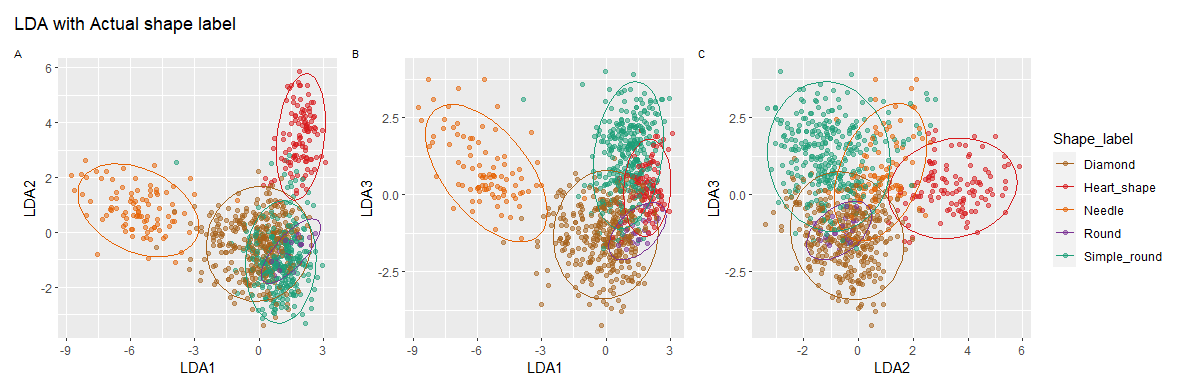}

  \medskip
  \includegraphics[width=0.95\textwidth]{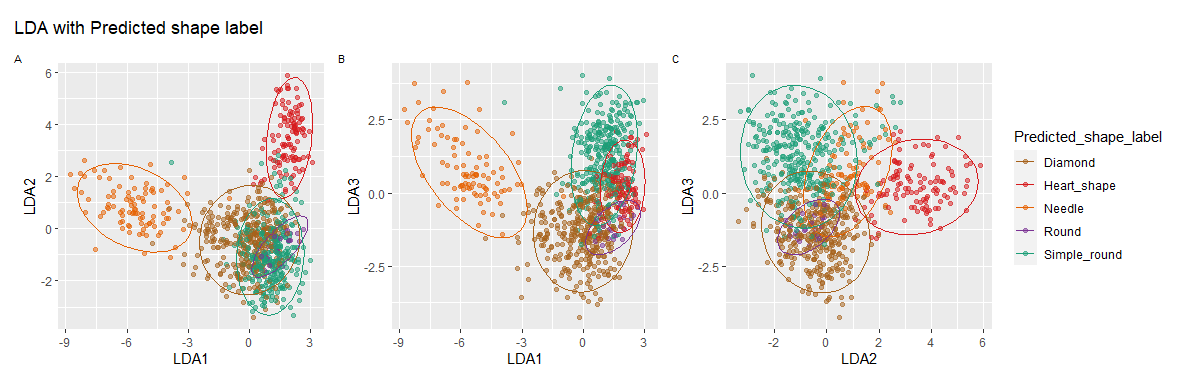}
  \caption {Needle-shaped leaves separate in every view; the other shapes overlap. The three pairwise LDA views of the Actual dataset: (a) actual shape labels; (b) predicted shape labels.}
  \label{fig:ldapairs}
\end{figure}

\begin{figure}[htbp]
  \centering
  \includegraphics[width=0.95\textwidth]{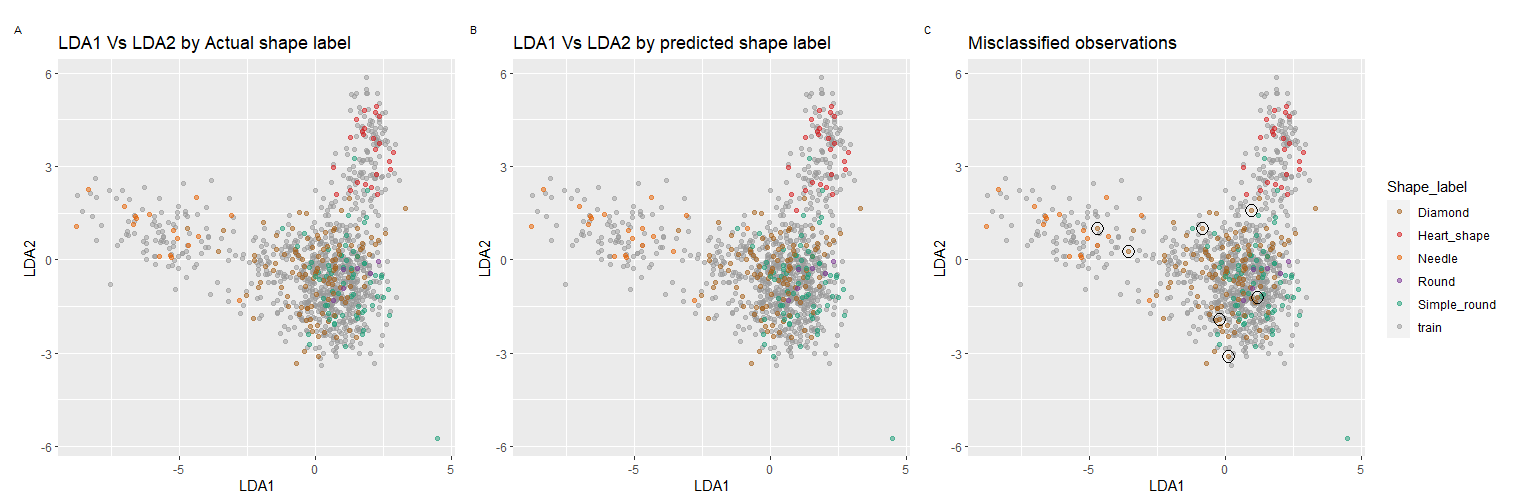}

  \medskip
  \includegraphics[width=0.95\textwidth]{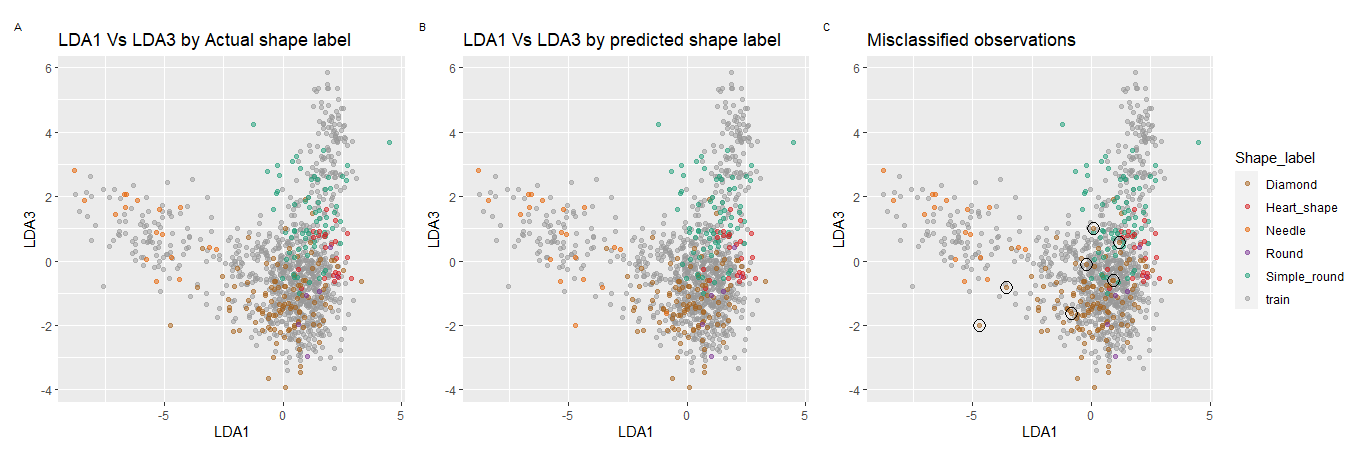}

  \medskip
  \includegraphics[width=0.95\textwidth]{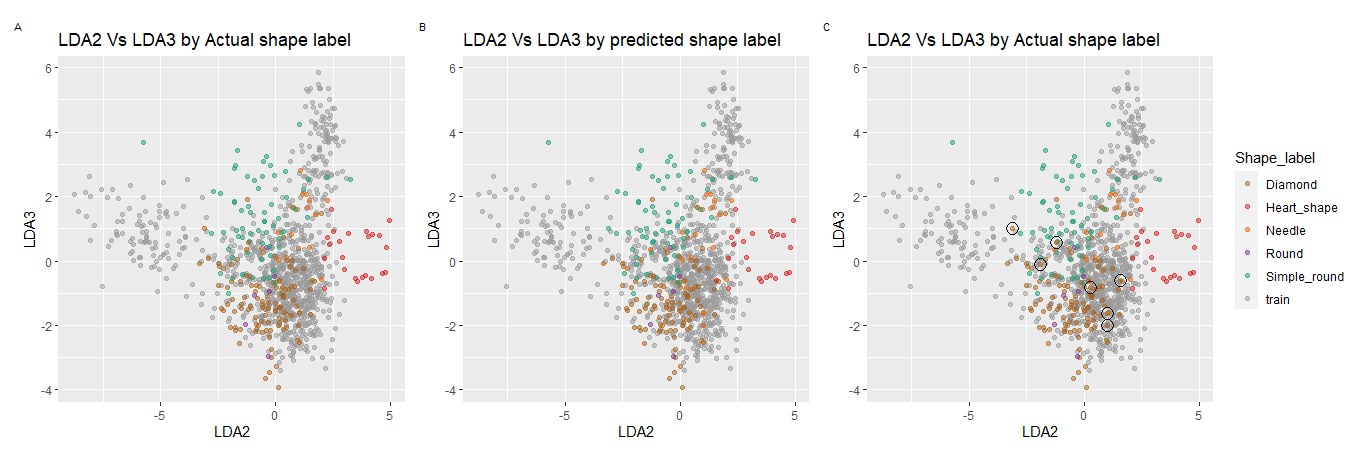}
  \caption{Seven leaves are misclassified as diamond in every view. Actual labels, predicted labels, and misclassified observations of the Actual dataset on (a) LDA1 vs.\ LDA2, (b) LDA1 vs.\ LDA3, and (c) LDA2 vs.\ LDA3.}
  \label{fig:misclass}
\end{figure}

\begin{figure}[htbp]
  \centering
  \includegraphics[width=0.9\textwidth]{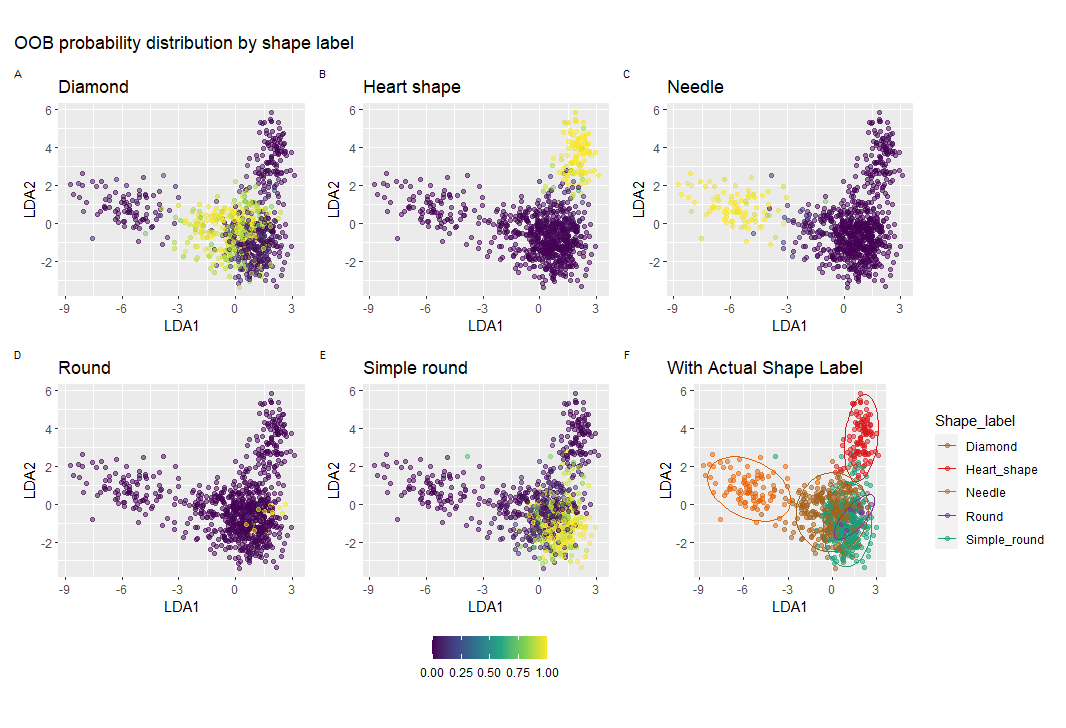}
  \caption{Random Forest is confident about every shape inside the training space. LDA1 vs.\ LDA2 colored by out of bag (OOB) probability for each shape label.}
  \label{fig:oob}
\end{figure}

\begin{figure}[htbp]
  \centering
  \includegraphics[width=0.48\textwidth]{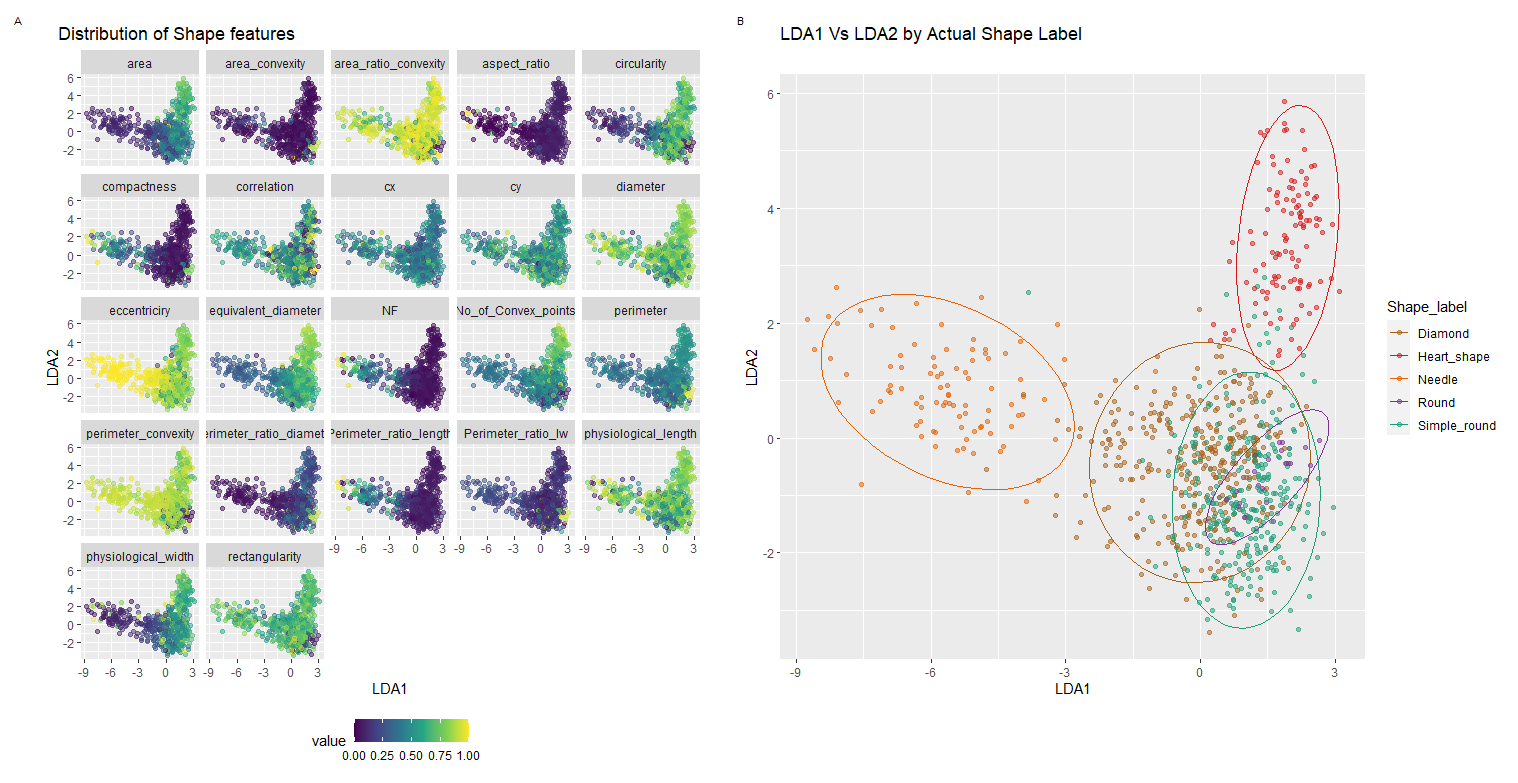}\hfill
  \includegraphics[width=0.48\textwidth]{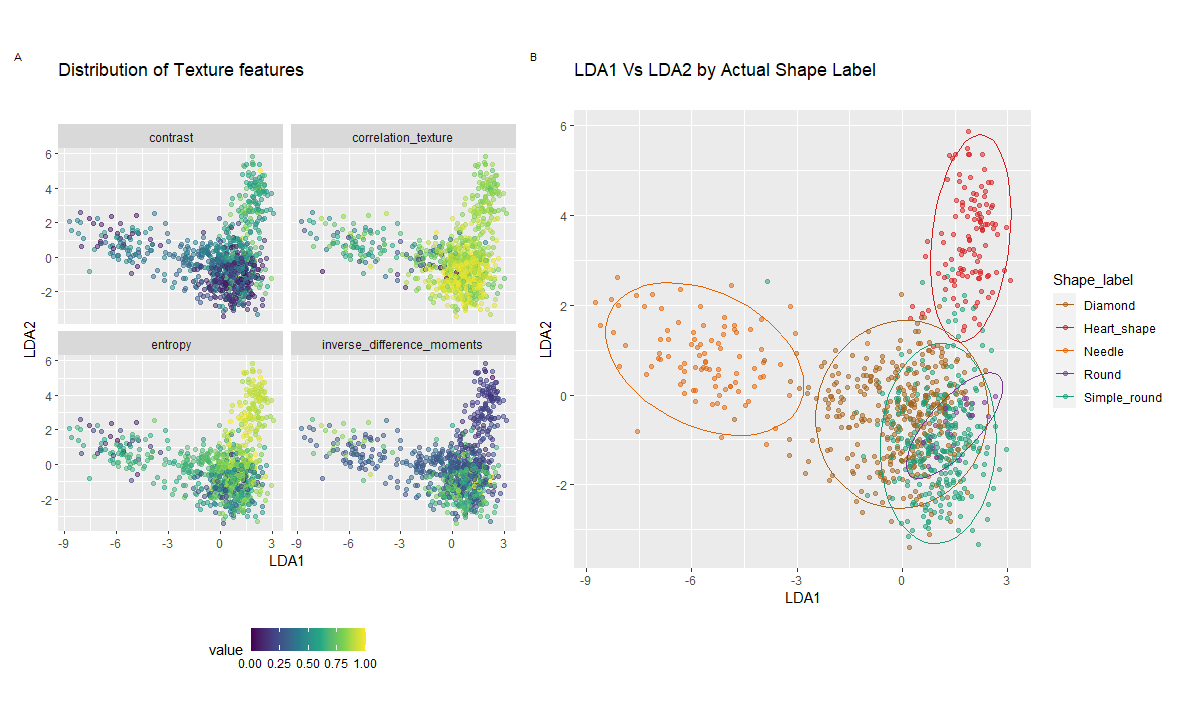}

  \medskip
  \includegraphics[width=0.48\textwidth]{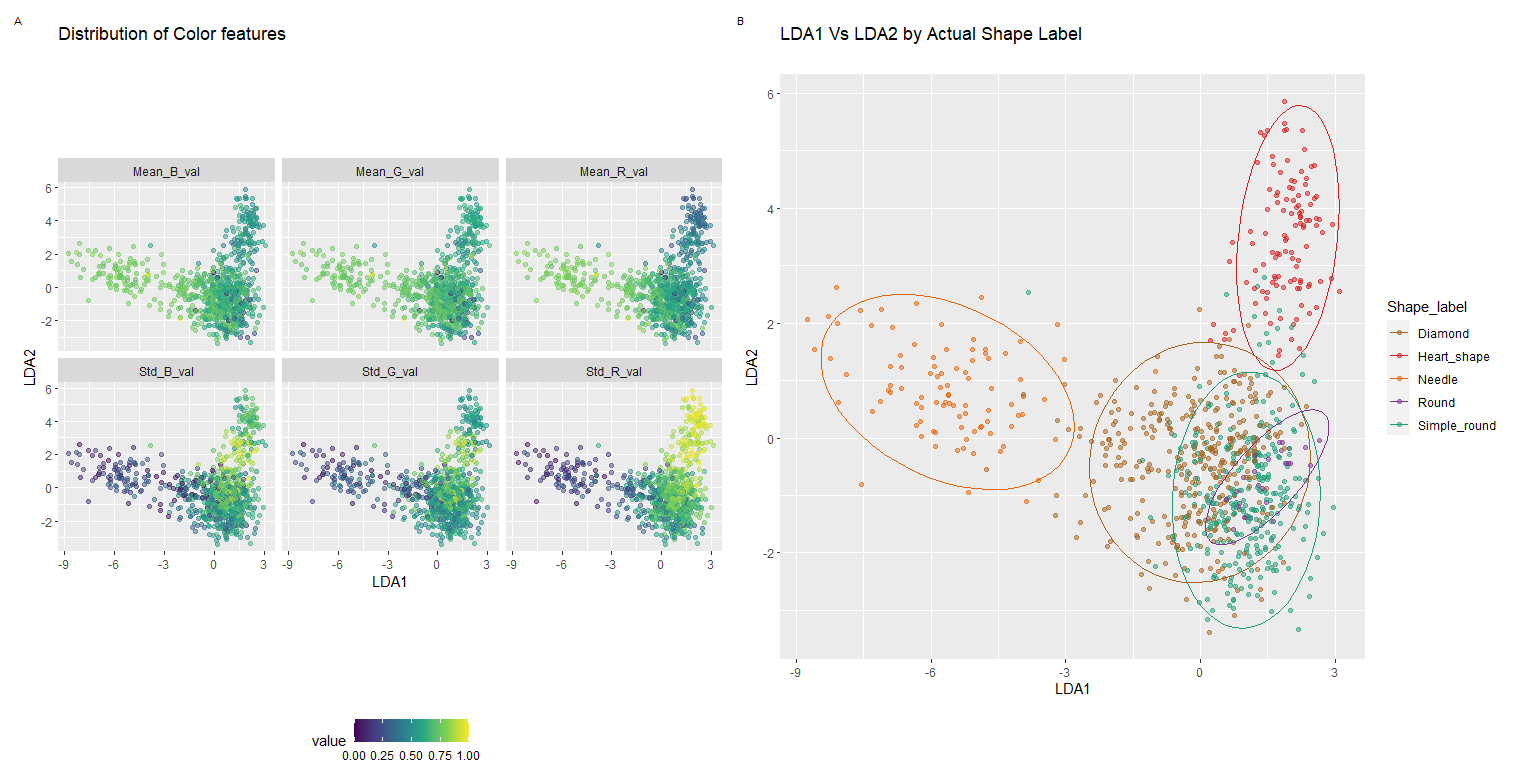}\hfill
  \includegraphics[width=0.48\textwidth]{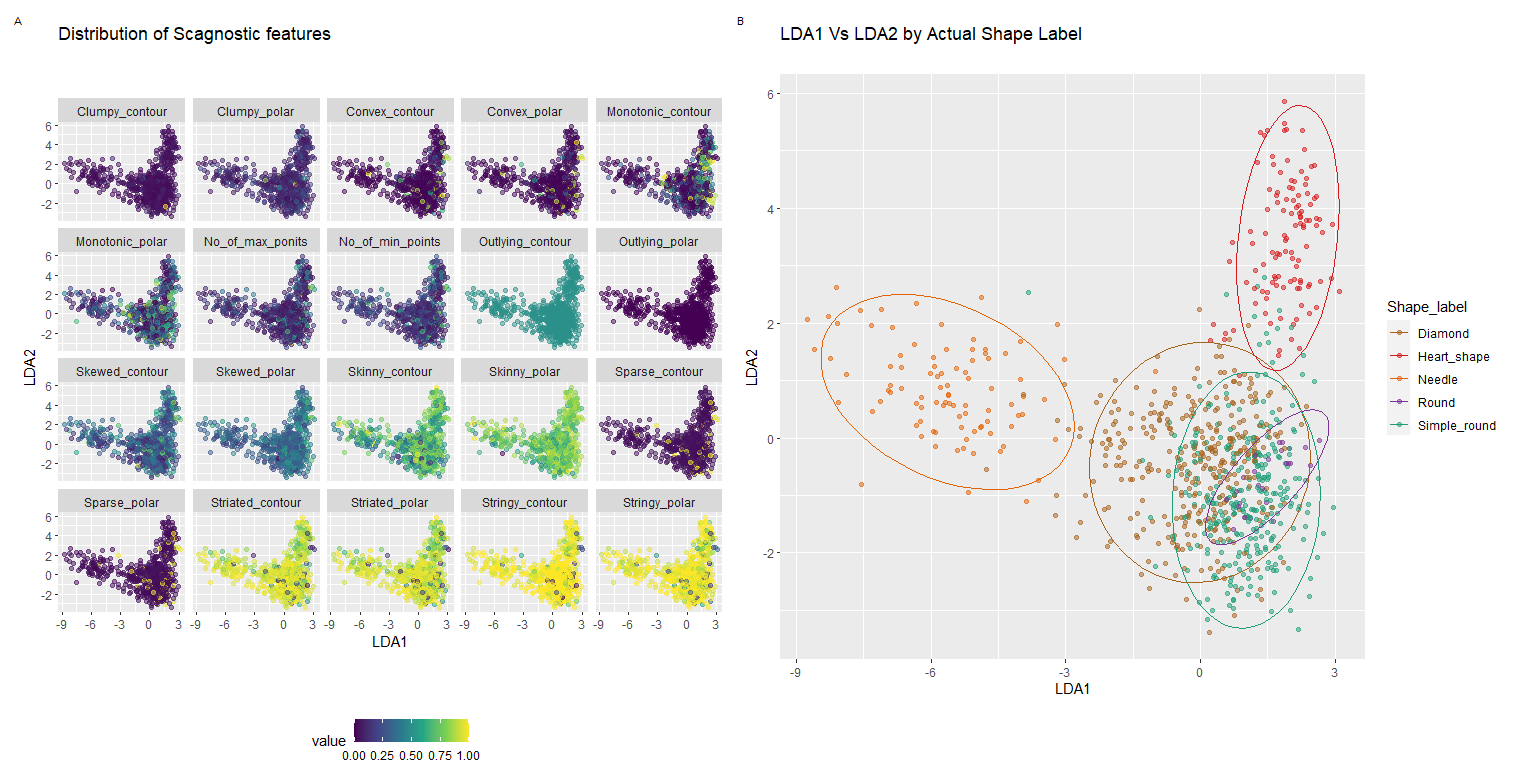}
  \caption{Each feature category separates the shape classes in a different way. LDA1 vs.\ LDA2 of the Actual dataset using only (a) shape, (b) texture, (c) color, or (d) scagnostic features.}
  \label{fig:featlda}
\end{figure}

Random Forest is often treated as a black box, so we used Linear Discriminant
Analysis (LDA) to look inside the trained model. LDA combines the many leaf
features into a few new axes (LDA1, LDA2, and LDA3) that best separate the shape
classes. Plotting leaves on these axes shows how well the shapes are separated and
where the model gets confused. We first look at the shape labels (Figures~\ref{fig:lda12}
to~\ref{fig:oob}), and then at the features that drive the separation
(Figure~\ref{fig:featlda} and Table~\ref{tab:featsummary}).

\hypertarget{training-and-test-both-are-in-actual-leaf-image-dataset}{%
\subsubsection{Training and Test Sets from the Actual Dataset}
\label{training-and-test-both-are-in-actual-leaf-image-dataset}}

All plots in this section use the Actual dataset. In the first three figures, the left panels
show the training set only, and the right panels add the test set. The top panels
use the actual shape labels, and the bottom panels use the labels predicted by
Random Forest.

On the first two axes (LDA1 and LDA2), needle- and heart-shaped leaves separate
cleanly, while diamond, round, and simple round leaves overlap (Figure~\ref{fig:lda12}a). Test leaves fall in the same space as
the training leaves (Figure~\ref{fig:lda12}b), with one exception: a single simple
round leaf lies outside the training space. Random Forest predicts it correctly
(Figure~\ref{fig:lda12}c,d).

On LDA1 and LDA3, needle-shaped leaves are again separated, and the other four
shapes overlap (Figure~\ref{fig:lda13}). Three test leaves lie outside the training space. Two are
simple-round and are predicted correctly. The third is actually a diamond leaf but
is predicted as a needle.

On LDA2 and LDA3, all shape labels overlap, and three test leaves fall outside the
training space in both the actual and predicted label views (Figure~\ref{fig:lda23}).

Putting the three views side by side (Figure~\ref{fig:ldapairs}) gives a clear summary.
Needle-shaped leaves are classified correctly in every view. Heart-shaped leaves are
classified correctly in the LDA1--LDA2 view. Diamond, round, and simple round leaves
overlap in all views, which is where most classification errors arise.

Figure~\ref{fig:misclass} marks the misclassified leaves in each view. Seven leaves are
misclassified as diamond in all three views. Of these, three are actually needle-
shaped, one is heart-shaped, and three are simple round.

Random Forest also gives a probability for each prediction, based on the trees that
did not use that leaf in training (the out of bag, or OOB, probability). Within the
training space, these probabilities reach 1 for every shape label, which confirms a
good separation (Figure~\ref{fig:oob}).

\hypertarget{features-that-separate-the-shapes}{%
\subsubsection{Features That Separate the Shapes}
\label{features-that-separate-the-shapes}}

The same LDA plots can be drawn for each feature category separately (Figure~\ref{fig:featlda}).
These plots show which features help to tell the shapes apart. Table~\ref{tab:featsummary}
summarizes what we found.

\begin{table}[htbp]
\centering
\caption{Features that help to identify each leaf shape (from Figure~\ref{fig:featlda}).}
\label{tab:featsummary}
\small
\begin{tabularx}{\textwidth}{lLLLL}
\toprule
Shape & Shape features & Texture & Color & Scagnostic \\
\midrule
Needle & Higher: eccentricity, compactness, narrow factor, perimeter ratio length. Lower: circularity, area, equivalent diameter, physiological length & -- & Lower standard deviation of red & -- \\
Heart & Higher number of convex points & Higher contrast, entropy, and correlation & Lower mean of blue, green, and red; higher standard deviation of red, green, and blue & -- \\
Round & Higher: area convexity, compactness, correlation, perimeter, perimeter ratio diameter, perimeter ratio length and width. Lower: area ratio convexity, number of convex points, perimeter convexity, rectangularity & Higher correlation & Higher standard deviation of blue and green & Higher monotonic contour and monotonic polar \\
Simple round & -- & Higher inverse difference moment and correlation & Higher standard deviation of blue and green & -- \\
Diamond & -- & -- & -- & -- \\
\bottomrule
\end{tabularx}
\end{table}

Shape features separate needle, heart, and round leaves best (Figure~\ref{fig:featlda}a).
Texture features mainly help to pick out heart leaves, and also split heart, round,
and simple round leaves from needle and diamond leaves (Figure~\ref{fig:featlda}b).
Color features identify heart leaves, which are darker than the other shapes, and
needle leaves, which vary least in red (Figure~\ref{fig:featlda}c). Scagnostic
features help mainly with round leaves (Figure~\ref{fig:featlda}d). No feature
category singles out diamond leaves, which fits with their overlap with other shapes
in the earlier plots. Striated and stringy scagnostics are high for all shapes, and
clumpy, convex, outlying (polar), and sparse scagnostics are low for all shapes, so
they do not help to tell the shapes apart.

\subsubsection{Visualization of Actual, Flavia, Swedish, and Kaggle Leaf Datasets}

\begin{figure}[htbp]
  \centering
  \includegraphics[width=0.30\textwidth]{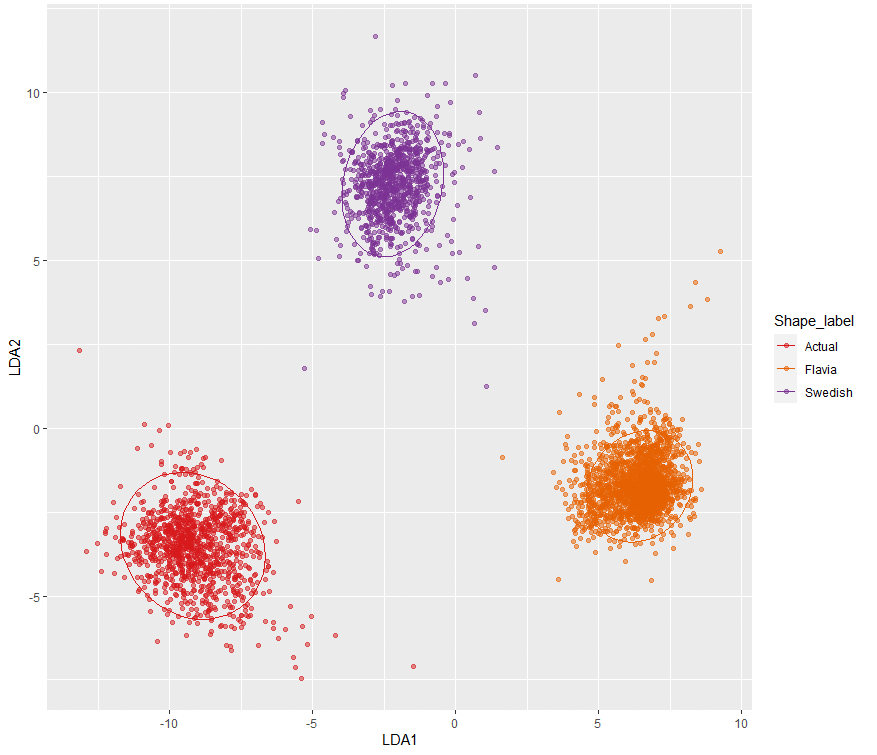}\hfill
  \includegraphics[width=0.30\textwidth]{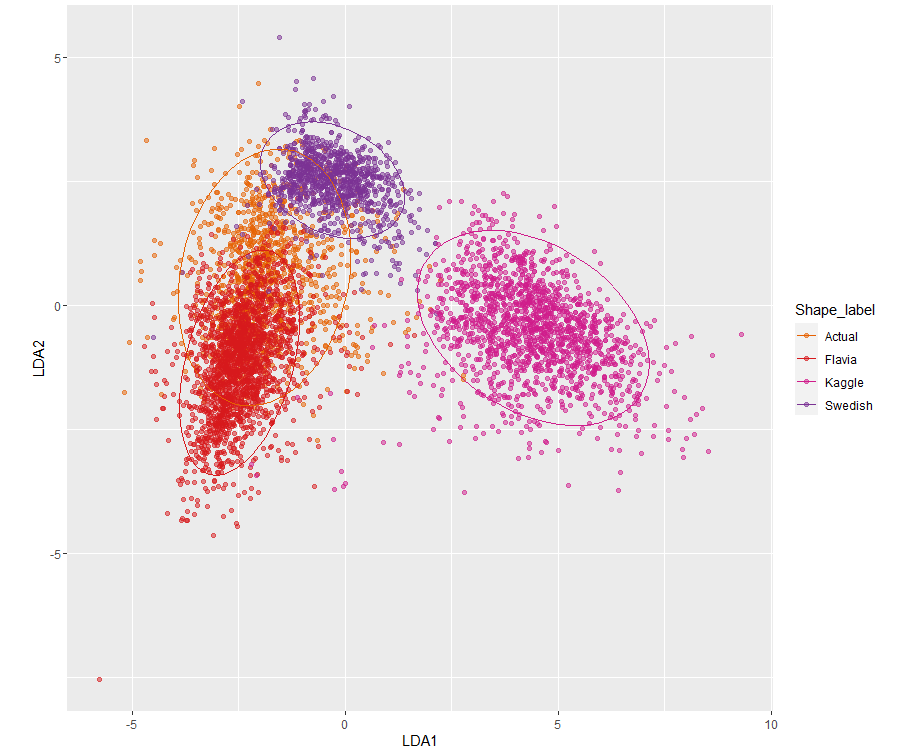}\hfill
  \includegraphics[width=0.30\textwidth]{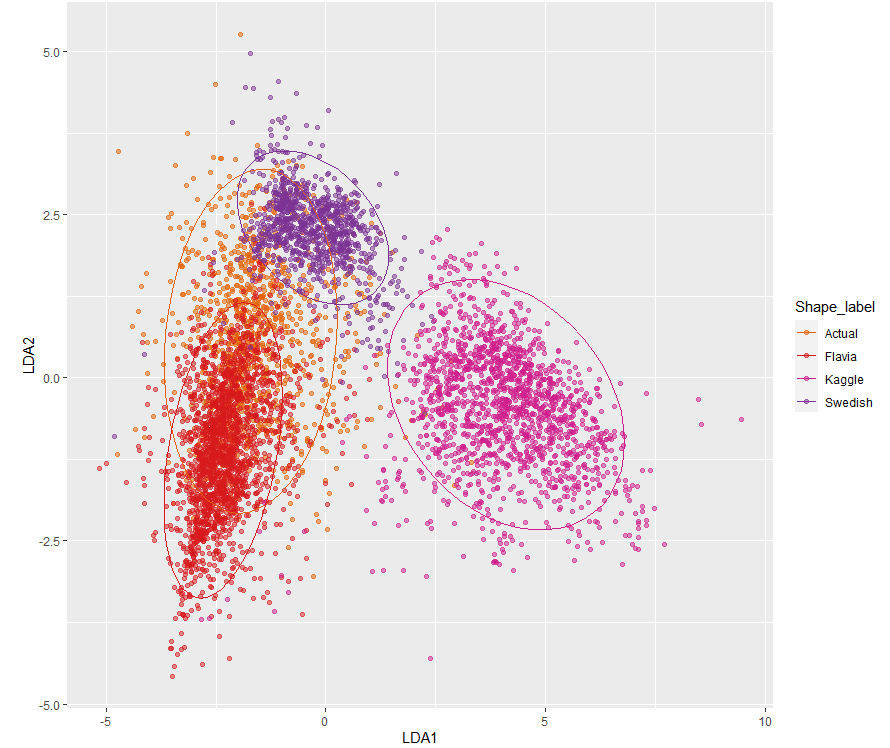}
  \caption{The Kaggle dataset lies apart from the other three datasets in LDA space. LDA1 vs.\ LDA2 of (a) the Actual, Flavia, and Swedish datasets using all feature categories, and (b, c) the Actual, Flavia, Swedish, and Kaggle datasets using shape features only.}
  \label{fig:40}
\end{figure}

We also used LDA to compare the four leaf datasets with one another. The Actual,
Flavia, and Swedish datasets contain color images, so we could extract all four
feature categories (shape, texture, color, and scagnostic), giving 52 features
in total. With all of these features, the three datasets project into overlapping
regions of LDA space (Figure~\ref{fig:40}a).

The Kaggle dataset contains only binary images, so only shape and scagnostic
features could be extracted for it. When we compare all four datasets using these
two feature categories (Figure~\ref{fig:40}b,c), the Kaggle leaves form a separate
region that does not overlap the other three, while Actual, Flavia, and Swedish
still overlap one another. This difference between Kaggle and the other datasets
may help to explain why accuracy was low when Kaggle was used in the cross dataset
experiment (Table~\ref{tab:10}).

\subsection{Feature Importance}
\label{sec:featimportance}

Variable importance shows how much each feature contributes to a model's
predictions, so it tells us which features matter most for classifying leaves.
We looked at feature importance at each level of the hierarchy, in the same order
as the classification: shape (H1), edge type (H2), and species (H3). All results
are for the Actual dataset.

\begin{figure}[htbp]
  \centering
  \includegraphics[width=0.95\textwidth]{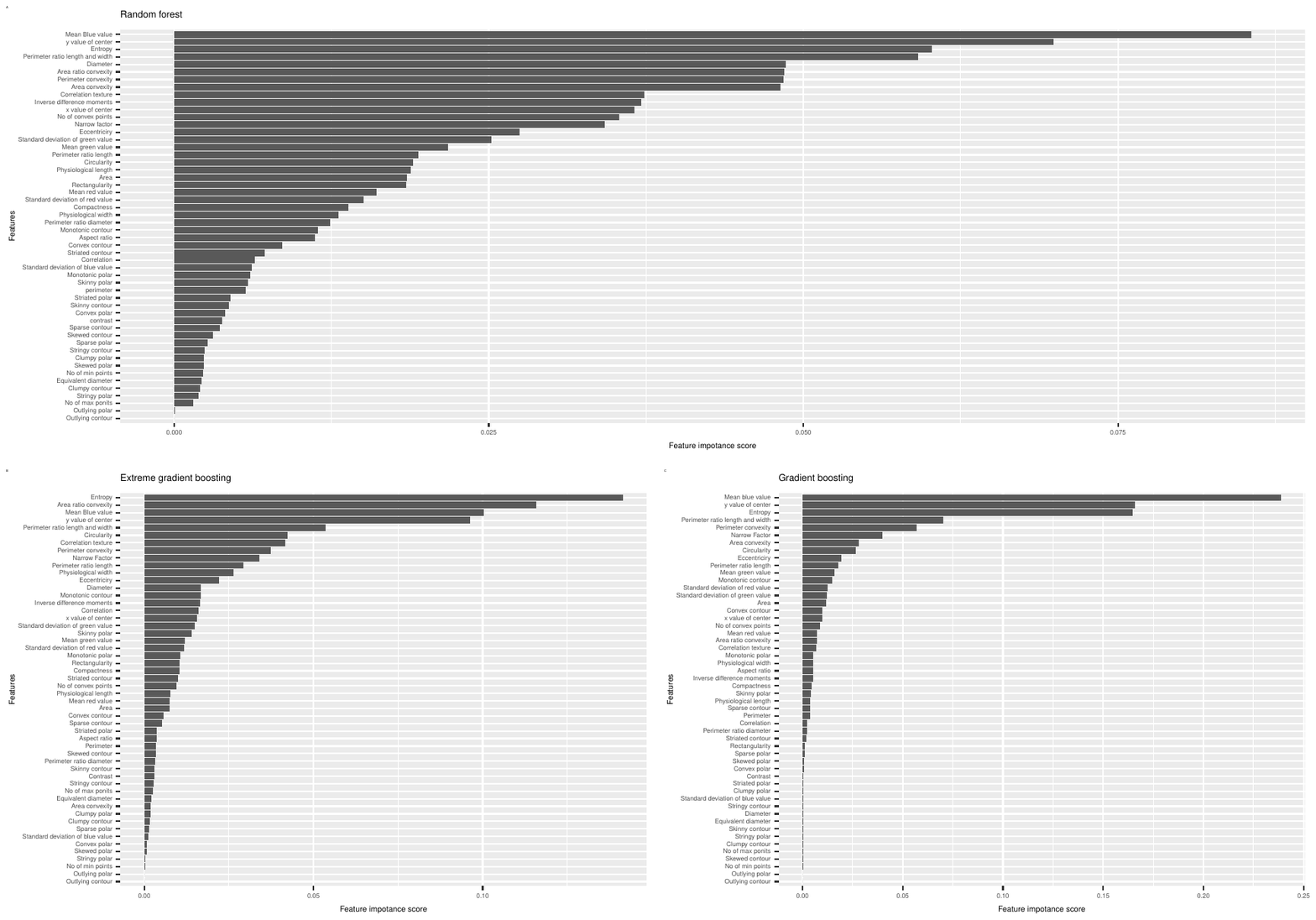}
  \caption{Shape-related features are the most important for classifying leaf shape. Feature importance in shape-wise (H1) classification of the Actual dataset: (A) Random Forest, (B) XGBoost, (C) Gradient Boosting.}
  \label{fig:impshape}
\end{figure}

At the first level, we compared the variable importance of Random Forest, XGBoost,
and Gradient Boosting (Figure~\ref{fig:impshape}). The most important features
are mainly shape features: the $x$ and $y$ values of the leaf center, the perimeter
ratio of length and width, diameter, area convexity, perimeter convexity, narrow
factor, area ratio convexity, physiological length, physiological width,
rectangularity, and eccentricity. Entropy is also among the most important
features.

\begin{figure}[htbp]
  \centering
  \includegraphics[width=0.95\textwidth]{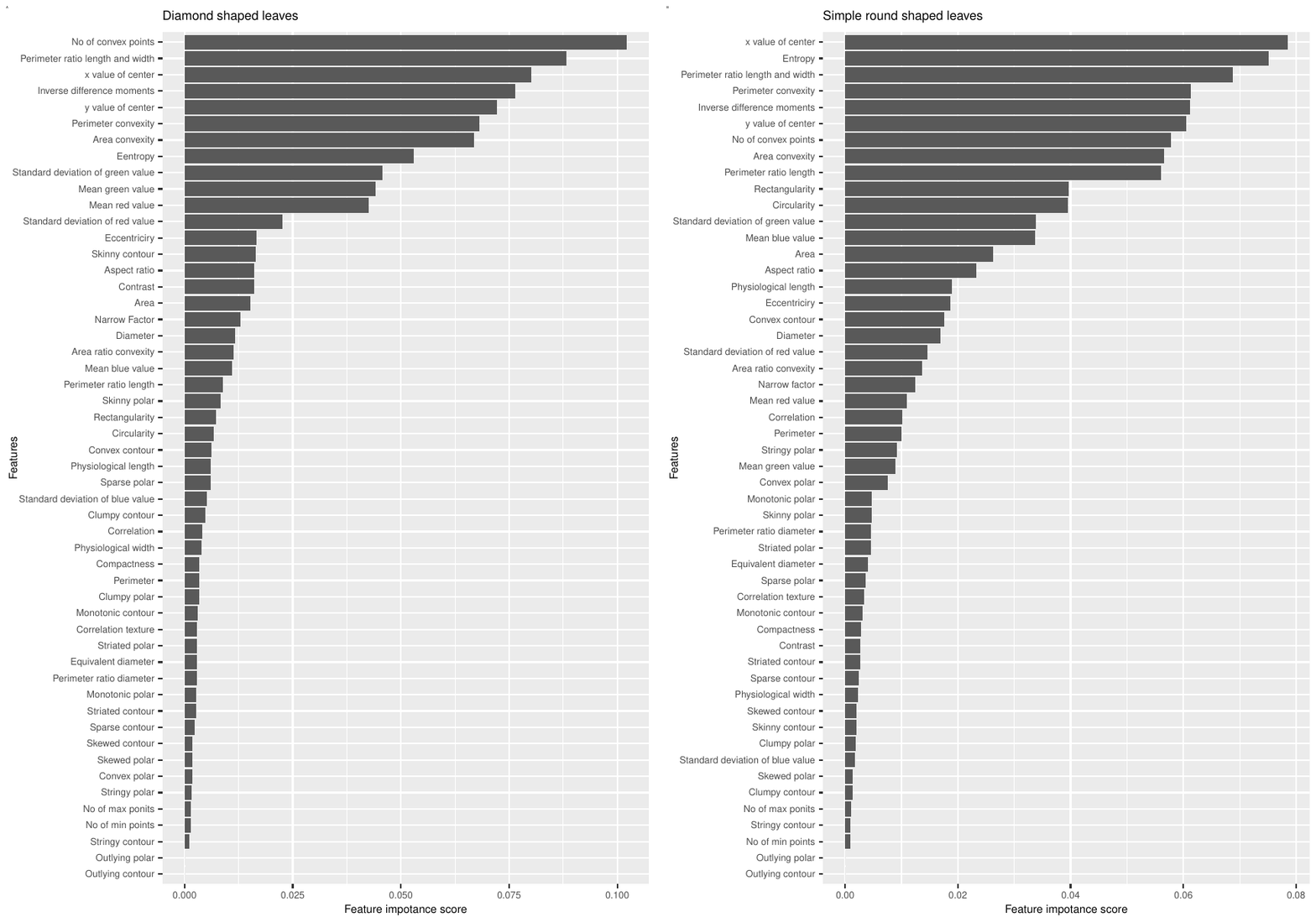}
  \caption{Feature importance in edge-wise (H2) classification of the Actual dataset: (A) diamond shaped leaves, (B) simple round shaped leaves.}
  \label{fig:impedge}
\end{figure}

For the lower levels, we used Random Forest. Figure~\ref{fig:impedge} shows the most important features for classifying edge types in diamond-shaped and simple round-shaped leaves. For diamond-shaped leaves, the number of convex points, perimeter ratio length and width, and x value of the center were the most important features. For simple round-shaped leaves, the x value of the center, entropy, and perimeter ratio length and width were the most important features. These results show that different features are important for identifying edge types in different leaf shapes.

\begin{figure}[htbp]
  \centering
  \includegraphics[width=0.95\textwidth]{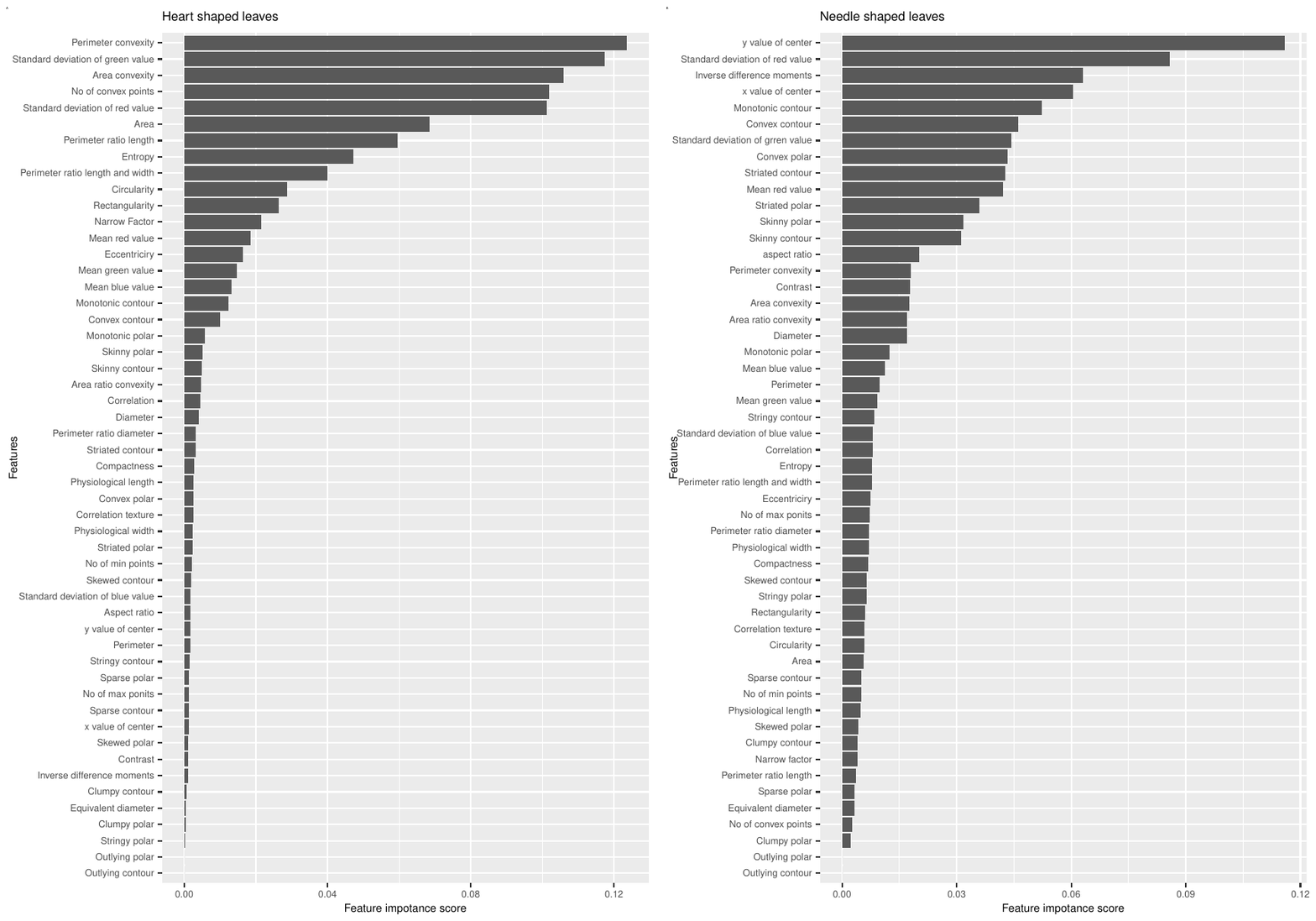}

  \medskip
  \includegraphics[width=0.48\textwidth]{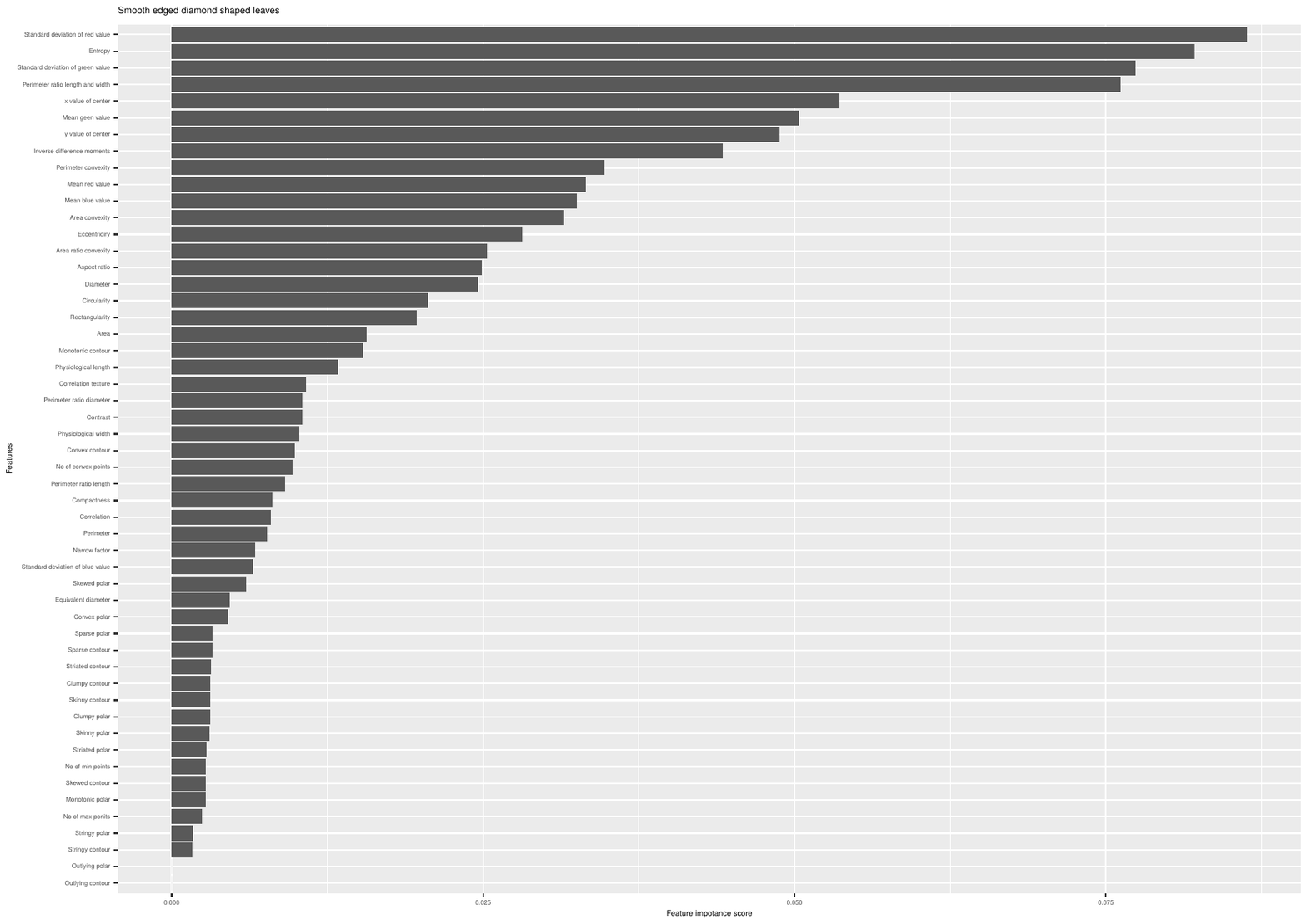}\hfill
  \includegraphics[width=0.48\textwidth]{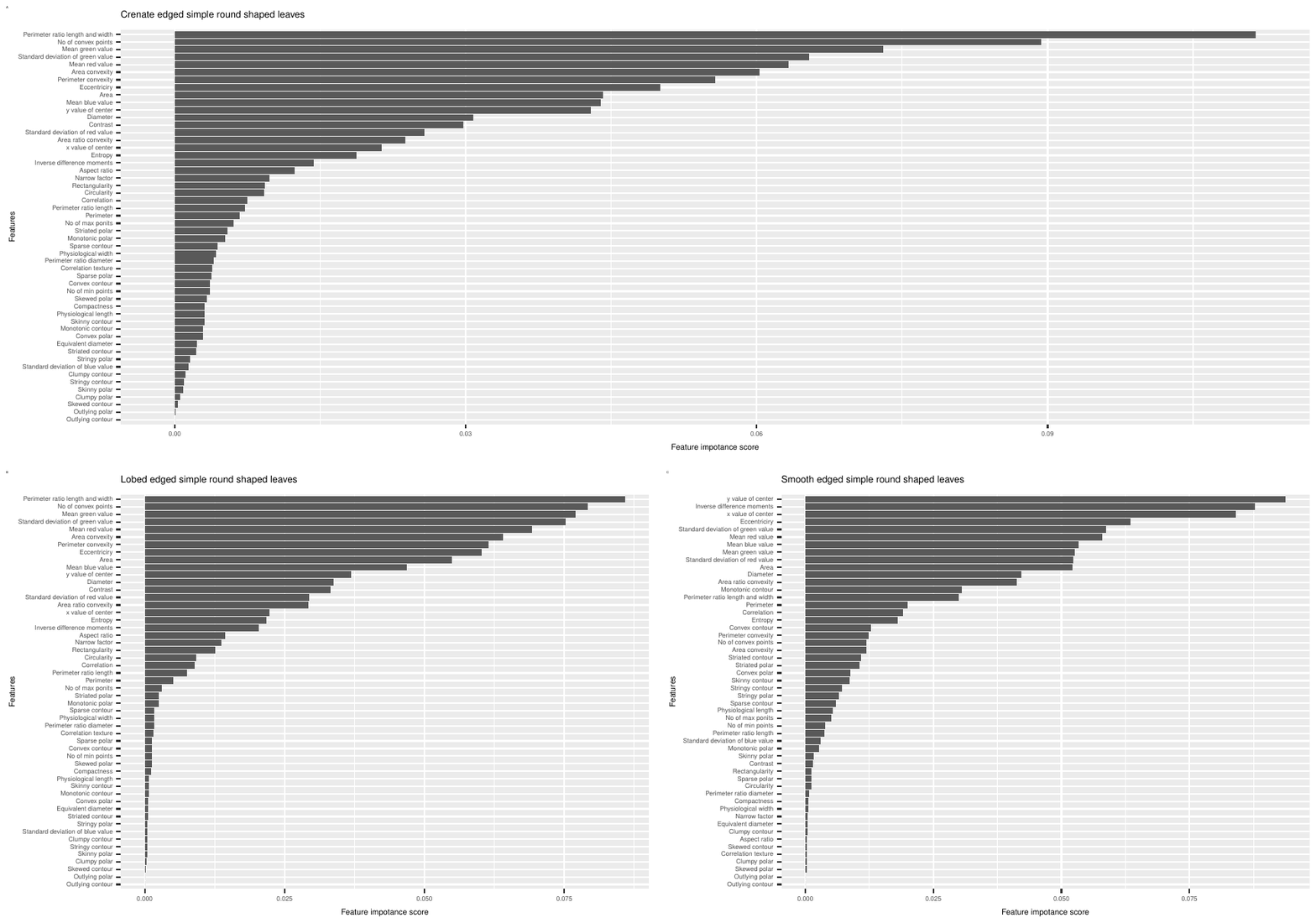}
  \caption{Scagnostic features matter most for classifying species within diamond and simple round shaped leaves. Feature importance in species-wise (H3) classification of the Actual dataset: heart shaped leaves (top left), needle shaped leaves (top right), smooth edged diamond-shaped leaves (bottom left), and simple round shaped leaves (bottom right).}
  \label{fig:impspecies}
\end{figure}

At the species level, Figure~\ref{fig:impspecies} shows the important features for
heart and needle shaped leaves (top), and for smooth edged diamond leaves and
simple round leaves (bottom). For the diamond and simple round leaves, scagnostic
features become more important: monotonic contour, convex polar, convex contour,
striated polar, striated contour, and skinny contour.

\section{Discussion}
\label{sec:discussion}

Identifying medicinal plant species automatically from leaf images has many
practical uses. In this study we built MEDIPI (MEDIcinal Plant Identification), a
classifier that works from ordinary leaf photographs. Leaf images carry rich
information about shape, edges, texture, and color, and we turned this information
into 52 interpretable features. Many earlier studies use shape, texture, and color
features only. We added a new scagnostics category, together with three new shape
features: correlation of the Cartesian contour, number of convex points, and number
of minimum and maximum points. Both the supervised and the unsupervised projections
(linear discriminant analysis and principal component analysis) show clear
separation between classes, so these features do carry useful information.

A second contribution is the data. Most existing work relies on datasets such as
Flavia and Swedish, which have few plant species and are too small to train a
reliable model on their own. No image database of Sri Lankan medicinal plants
existed, so we built one and released it in the open-source R package
\texttt{MedLEA} (\url{https://CRAN.R-project.org/package=MedLEA}). We first studied
the leaves in the medicinal plant repository of the Barberyn Ayurveda resort and the
University of Ruhuna (\url{http://www.instituteofayurveda.org/plants/}) and
identified five main shapes and four main edge types. The images were then collected
in a simple way that needs no expert knowledge: a normal smartphone, no flash to
avoid shadows, and the leaf placed at the center of a white sheet of paper.

The results support three main findings. First, a hierarchical Random Forest
classifier worked best. Random Forest gave the highest accuracy of the three
tree-ensemble algorithms we tried, and it also performed better than the benchmark
algorithms (Table~\ref{tab:12}). Classifying first by shape, then by edge type, and
then by species let the model use different features at each step. Shape features
such as the leaf center coordinates, perimeter ratio of length and width, diameter,
convexity measures, narrow factor, physiological length and width, rectangularity,
and eccentricity were the most important for the first level, while scagnostic
features such as monotonic contour, convex polar, and striated contour mattered more
for identifying species at the bottom level (Section~\ref{sec:featimportance}).

Second, shape features alone are not enough. When we compared models trained on shape
features only with models trained on all feature categories, the all feature models
were more accurate. For example, Random Forest on the Actual dataset reached 97\%
shape level accuracy with all features but 90\% with shape features only
(Tables~\ref{tab:5} and~\ref{tab:9}). Combining feature categories within a
hierarchical structure therefore gave the strongest results.

Third, the model is not a complete black box. The LDA plots show how the trained
Random Forest separates the shape classes: needle shaped leaves are separated in
every view, while diamond, round, and simple round leaves overlap, which is where
most of the errors occur (Section~\ref{visualization-of-algorithm-performance}).

Our results also show some limitations. Accuracy was high only when the training and
test sets came from the same dataset (Experiment~1). When we trained on one dataset and
tested on another, accuracy dropped sharply (Experiment~2, Tables~\ref{tab:10}
and~\ref{tab:11}). One reason may be that the datasets differ in how the images were
collected: for example, the Kaggle images are binary only and lie apart from the
other datasets in LDA space. Cross dataset testing was done only at the shape level,
so we do not yet know how well the edge and species levels would transfer. In
addition, the Actual dataset covers only a limited set of species from Sri Lanka, and
some species were classified less well (for example, a few smooth edged diamond
leaves in Experiment~1).

\section{Conclusions}
\label{sec:conclusions}

We presented MEDIPI, a hierarchical, feature-based statistical machine learning
method for classifying medicinal plant leaves, together with MedLEA, an open R
package of leaf images and morphological data from Sri Lanka. Using 52 interpretable
shape, texture, color, and scagnostic features, a hierarchical Random Forest gave
the best performance among the algorithms we compared, and linear discriminant
analysis helped us see how the model separates the leaf shapes. Shape features alone
were not enough; combining feature categories within a hierarchical structure worked
best. In future work, MEDIPI could be extended to more, and more diverse, medicinal
species, tested on images from other sources to improve cross dataset accuracy, and
compared directly with deep-learning methods to measure the trade-off between
accuracy and interpretability.

\section*{Author Contributions}
Conceptualization, J.P.G. and T.S.T.; methodology, J.P.G. and T.S.T.; software, J.P.G.; validation, J.P.G. and T.S.T.; formal analysis, J.P.G.; investigation, J.P.G.; resources, J.P.G. and T.S.T.; data curation, J.P.G.; writing -- original draft preparation, J.P.G. and T.S.T.; writing -- review and editing, J.P.G. and T.S.T.; visualization, J.P.G. and T.S.T.; supervision, T.S.T.; project administration, T.S.T.

\section*{Funding}
This research received no external funding.

\section*{Data Availability Statement}
The data presented in this study are openly available in the R package MedLEA at \url{https://CRAN.R-project.org/package=MedLEA}. This repository contains the leaf images and morphological data collected for this study, along with the labeling scheme used to build the hierarchical classifier. The Flavia, Swedish, and Kaggle leaf image datasets analyzed in this study are third party datasets used under their original terms of release; restrictions may apply to their redistribution, and they should be obtained directly from their original sources.

\section*{Acknowledgments}
Not applicable.

\section*{Conflicts of Interest}
The authors declare no conflicts of interest.

\section*{Abbreviations}
The following abbreviations are used in this manuscript:\\[4pt]
\begin{tabular}{@{}ll}
IUCN & International Union for Conservation of Nature\\
MEDIPI & MEDIcinal Plant Identification\\
LDA & Linear Discriminant Analysis\\
RF & Random Forest\\
XGBoost & Extreme Gradient Boosting\\
KNN & $k$-Nearest Neighbor\\
SVM & Support Vector Machine\\
PNN & Probabilistic Neural Network\\
ANN & Artificial Neural Network\\
CNN & Convolutional Neural Network
\end{tabular}

\bibliographystyle{unsrtnat}
\bibliography{references}

\end{document}